\documentclass[12pt, draftclsnofoot, onecolumn]{IEEEtran}

\usepackage{amssymb}
\usepackage{amsmath}
\usepackage{cite}
\usepackage{url}
\usepackage{xcolor}
\usepackage{cite,graphicx,amsmath,amssymb}
\usepackage{subfigure}
\usepackage{citesort}
\usepackage{fancyhdr}
\usepackage{mdwmath}
\usepackage{mdwtab}
\usepackage{caption}
\usepackage{amsthm}
\usepackage{algorithm}
\usepackage{algorithmic}
\usepackage{graphicx} 
\usepackage{epstopdf}
\usepackage{setspace}
\usepackage{tikz}

\newtheorem{remark}{Remark}
\newtheorem{theorem}{Theorem}

\newtheorem{lemma}{Lemma}

\newtheorem{corollary}{Corollary}

\makeatletter
\def\ScaleIfNeeded{%
\ifdim\Gin@nat@width>\linewidth \linewidth \else \Gin@nat@width
\fi } \makeatother

\begin{document}

\title{Path Design and Resource Management for NOMA enhanced Indoor Intelligent Robots}

\author{Ruikang~Zhong,~\IEEEmembership{Student Member,~IEEE,}
Xiao~Liu,~\IEEEmembership{Student Member,~IEEE,}
Yuanwei~Liu,~\IEEEmembership{Senior Member,~IEEE,}
Yue~Chen,~\IEEEmembership{Senior Member,~IEEE,}
Xianbin~Wang,~\IEEEmembership{Fellow,~IEEE,}

\thanks{Ruikang~Zhong, Xiao~Liu, Yuanwei~Liu, and Yue~Chen are with the Queen Mary University of London, London E1
4NS, U.K. (e-mail: r.zhong@qmul.ac.uk; x.liu@qmul.ac.uk; yuanwei.liu@qmul.ac.uk; yue.chen@qmul.ac.uk).

Xianbin Wang is with Department of Electrical and Computer Engineering, Western University, London, ON N6A5B9, Canada (e-mail: xianbin.wang@uwo.ca).

}

}

\maketitle
\begin{abstract}
A communication enabled indoor intelligent robots (IRs) service framework is proposed, where non-orthogonal multiple access (NOMA) technique is adopted to enable highly reliable communications. In cooperation with the ultramodern indoor channel model recently proposed by the International Telecommunication Union (ITU), the Lego modeling method is proposed, which can deterministically describe the indoor layout and channel state in order to construct the radio map. The investigated radio map is invoked as a virtual environment to train the reinforcement learning agent, which can save training time and hardware costs.
Build on the proposed communication model, motions of IRs who need to reach designated mission destinations and their corresponding down-link power allocation policy are jointly optimized to maximize the mission efficiency and communication reliability of IRs.
In an effort to solve this optimization problem, a novel reinforcement learning approach named deep transfer deterministic policy gradient (DT-DPG) algorithm is proposed.  Our simulation results demonstrate that 1) With the aid of NOMA techniques, the communication reliability of IRs is effectively improved; 2) The radio map is qualified to be a virtual training environment, and its statistical channel state information improves training efficiency by about 30\%; 3) The proposed DT-DPG algorithm is superior to the conventional deep deterministic policy gradient (DDPG) algorithm in terms of optimization performance, training time, and anti-local optimum ability.

\end{abstract}

\begin{keywords}
Indoor path design, intelligent robot, non-orthogonal multiple access, radio map, reinforcement learning
\end{keywords}

\section{Introduction}

The explosive development of robotics and artificial intelligence technologies have changed, are changing and will continue to transform human lives. In recent years, intelligent robots (IRs) are proven competent to provide a variety of services, such as security monitoring, sanitation, and travel guides \cite{tan2019one}.  New various services offered by IRs require a large amount of communication, computation and data resources, which are not necessarily provided locally \cite{lindhe2012communication}. Therefore, in order to alleviate the requirements on local hardware resources, particularly computing power and memory, developing communication-empowered robots is regarded as a cost-effective solution~\cite{galambos2020cloud}.  Hence, we propose a communication enhanced robot service model in this paper, which maximally ensures the communication quality for scenarios such as robot engaged guiding, security, and etc.

Specifically, we take a squad of IRs providing shopping guide services as an example to illustrate the IR service model. Contemporary shopping guide services are generally completed by a human employee with a portable communication device. With assistances of the device, such as an internet-connected iPad, the salesclerk can display relevant videos, pictures or advertisements to the relevant customers. Simultaneously, the salesclerk also has to lead the customer to a destination such as the display racks in the store. In our IR service model, the two responsibilities of leading and presenting can be borne by one IR concurrently. Therefore, the IR needs to choose a time-saving path to the chosen destination and maintain a qualified wireless link with the core network in order to obtain the required information to display in real-time during the moving process. Meanwhile, considering the user capacity required by multiple IRs and the variable indoor channels state, the non-orthogonal multiple access (NOMA) technique is invoked due to its superiority in spectrum efficiency, connectivity, and user fairness \cite{NOMA5G.yuanwei}.

\subsection{The State of the Art}

\subsubsection{Path Planing and Radio Map}
In the early stage of research on robots' path planning, the common research aim was to find solutions to reach the destination without collision in a maze or an obstacle environment, and different algorithms were proposed, including the particle swarm optimization~\cite{zhang2013robot}, artificial bee colony algorithm \cite{bhattacharjee2011multi} and the regression search algorithm \cite{li2012efficient}.
Due to the communication demands of IRs, robot motion planning and communication are often considered simultaneously in a number of existing literatures. For example, in \cite{ghaffarkhah2012optimal}, the robots have to reach multiple locations successively and maintain communication with the base station, and a corresponding energy consumption minimization algorithm was proposed. The author of \cite{ghaffarkhah2011communication} proposed a motion design approach with a consideration of communication to maintain the link between the robot and base stations.

Employing radio maps is widely accepted as an important approach for indoor localization \cite{huang2019online} and network security~\cite{utkovski2019learning}. Several novel studies have proved that with the assistance of a radio map, the unmanned aerial vehicle (UAV) can obtain credible navigation and trajectories, which can effectively reduce the communication outage of the UAV \cite{zhang2019radio,liu2020machine}. Recently, as an intersection of the radio map and robotics, the author of \cite{mu2020intelligent} employed the channel power gain map and the intelligent reflecting surface (IRS) to provide a higher probability of line-of-sight (LoS) propagation and optimized trajectory for indoor robots. However, the influence of external interferences was ignored in this study, which deserves further exploration.


\subsubsection{Non-orthogonal Multiple Access in Robotic Networks}

Although the existing literatures do not specifically investigate the application of NOMA in robotic communications, IR is often considered as a component of Internet of Things (IoT) systems or smart devices, and NOMA techniques are considered to be suitable for machine communications since the huge number of devices requires enormous capacity \cite{budhiraja2019tactile,shirvanimoghaddam2017massive}. Furthermore, NOMA techniques are regarded as one of the candidate technologies for the IoT devices and vehicle communications since it has superior spectral efficiency and user fairness as well \cite{Yuanwei.1}. The author of \cite{zhang2017downlink} developed the analytical frameworks of up-link and down-link NOMA in a dense wireless network and proved the NOMA gain in terms of achievable rate and outage probability. Due to the high sensitivity on power, the power control policy of the NOMA enhanced system is a kernel of the optimization. Therefore, a series of related studies were proposed to optimize the delay \cite{zhai2019delay}, outage probability \cite{yang2016general}, and energy efficiency \cite{zhai2018energy} of the NOMA enhanced communication system.

\subsubsection{Reinforcement Learning (RL) in Robotic Control and Communications}

Invoking RL algorithms to control robots has achieved several remarkable successful cases \cite{zhang2020continuous,lobos2018visual}. In~\cite{kakish2020using}, the authors applied Q-learning and State-Action-Reward-State-Action (SARSA) algorithms to plan motions for a swarm of robots so that they can be deployed to a user-defined target distribution timely. The author of \cite{lakshmanan2020complete} adopted a deep reinforcement learning (DRL) algorithm with an actor-critic structure to optimize a complete coverage path for the tiling robot with minimal energy cost. In addition to robots, RL algorithms have been used to design the trajectory for other kinds of craft, such as vehicles \cite{james2019online,liu2020enhancing} and UAVs \cite{liu2018energy,liu2019trajectory,chen2019liquid,9140367}. From a perspective of the communication, since RL algorithms have a prominent ability to solve non-convex problems~\cite{liu2020}, it manifests extraordinary potential in wireless network optimization \cite{qian2019survey,jingjingwang1}. The author of \cite{yang2020deep} optimized the reflecting beamforming of an IRS by an RL approach and their results suggest that the participation of the RL significantly improves the secrecy and quality of service (QoS) satisfaction probability of the system. In \cite{zhao2019deep}, a dueling double deep Q-network algorithm was proposed to optimize the user association and resource allocation to maximize the long-term performance of the cellular network. Specifically for the NOMA system, the author of \cite{he2019joint} proposed a DRL algorithm to optimally assign channels and allocates power to maximize the sum rate and spectral efficiency of the system.

\subsection{Motivations and Contributions}

The research contributions mentioned above have provided valuable insights on robotic control and communications. However, in contrast to stationary IoT devices, such as wireless sensors in industry, facilities and electrical appliances, the mobility of robots and the location sensitivity of the indoor channels make problems such as the resource allocation highly dynamic, and DRL is considered to be an effective methodology for solving this kind of problems \cite{chen2019artificial,jingjingwang2}. Moreover, the obstacles and occlusions in the complex indoor environment are not likely to be functionally analytic, which poses a challenge to the path optimization problem since the control safety of IRs has to be guaranteed. As a consequence of above mentioned two reasons, we propose a DRL algorithm to jointly optimize trajectories and the power allocation policy of IRs. Exceeding the existing research contributions on indoor robot communications, we have the following three new contributions:

\begin{itemize}

\item An indoor IR service framework is proposed, where IRs need to find out efficient and communication-reliable trajectories to the designated destinations with an awareness of the channel quality provided by the radio map. Meanwhile, NOMA techniques are invoked to rescue the IR who is under poorer channel conditions. Based on the proposed communication model, we formulate the optimization problem to maximize the mission efficiency and communication reliability by jointly optimizing the motions and the power allocation of the IRs.

\item Built on the proposed NOMA communication framework, we develop a novel deep transfer deterministic policy gradient (DT-DPG) algorithm to solve the formulated problem. This algorithm is originally proposed for the first time. To the edge of our knowledge, no similar algorithm has been discovered yet in the field of communication. The proposed algorithm is capable of solving multi-objective optimization or optimization problems with preconditions. By the engagement of the transfer learning, the proposed DT-DPG algorithm overcomes the problem that the conventional deep deterministic policy gradient (DDPG) algorithm is likely to tramp into the local optimum.

\item We employ the radio map with the propagation knowledge of the indoor environment to train the RL agent. Correspondingly, we proposed a modeling approach for indoor furnishings to accurately depict the complex indoor occlusion state and build the radio map. The digital radio map can train the agent without the hardware consumption and physical venue expenditure. Our simulation results demonstrate that with the statistical channel state information provided by the radio map, the agent training is more effective and efficient.

\end{itemize}

\subsection{Organizations}

Section \uppercase\expandafter{\romannumeral2} illustrates the model of the NOMA enhanced IR system and the problem formulation. The function and construction of the indoor radio map are described in Section \uppercase\expandafter{\romannumeral3}. Section \uppercase\expandafter{\romannumeral4} presents the proposed DT-DPG algorithm for jointly optimizing the motions and the power allocation of IRs. Meanwhile, Section \uppercase\expandafter{\romannumeral4} also analyzes the complexity and convergence of the proposed DT-DPG algorithm. The simulation results are displayed and analyzed in Section \uppercase\expandafter{\romannumeral5}. At last, the conclusion is summarized in Section \uppercase\expandafter{\romannumeral6}.

\section{System Model}

\subsection{System Description and Assumption}

\begin{figure*}[t!] 
\centering 
\includegraphics[width=1\textwidth]{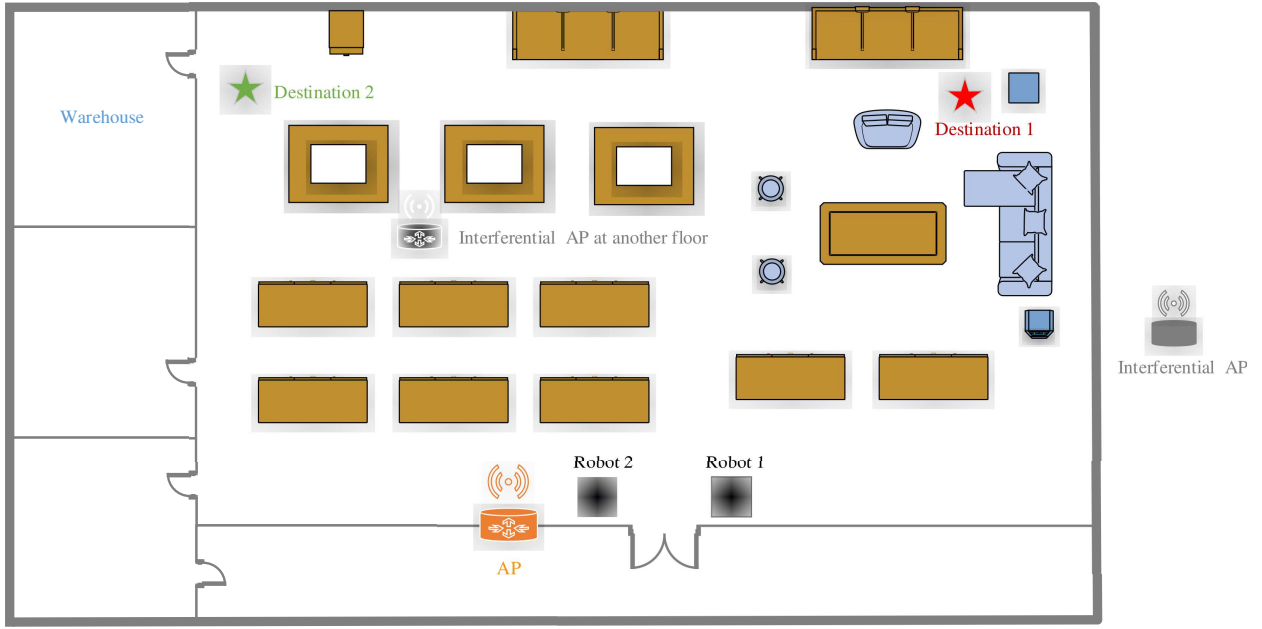} 
\caption{NOMA enhanced of indoor robots communication and path plan} 
\label{Fig.main2} 
\end{figure*}

We consider an indoor shopping mall scenario as an example where multiple IR servants are associated with a wireless access point (AP) as illustrated in Fig ~\ref{Fig.main2}. IRs in this service squad are responsible for receiving customers at the entrance and leading them to designated checkpoints, such as the waiting area or a designated product. As aforementioned, during the movement, the IR has to obtain the desired information of customers such as
product videos or advertisements via the Internet connection. Thus, the communication quality between IRs and the AP has to be guaranteed. However, as the IR continues to move, it may lose qualified links due to the complex multi-path effects of the indoor environment and potential external interference. Therefore, we propose a novel down-link model that invokes the knowledge of radio map to train the IR to minimize the task period and outage duration.

In this model, We denote the set of IRs as $u \in \mathbb{U}=\{1,2,3...U\}$ and we assume that each IR is equipped with a single antenna and employs NOMA techniques. IRs are associated with an authorized AP which denoted as $\boldsymbol{m}$ and the other interfering APs are denoted as $m \in \mathbb{M}=\{1,2,3...M\}$. Without loss of generality, we assume that all IRs associated with the AP are in one NOMA cluster, and multiple orthogonal resource blocks can be employed at the AP side to serve further user clusters in practice. We also assume that the surrounding radio environment does not change drastically, so the radio map can be regarded as accurate and valid during the mission period.

\subsection{Propagation Model}

The indoor path loss model for frequency range 300MHz to 450GHz has been proposed in ITU recommendation in 2019 ~\cite{ITUR}. This novel indoor channel model points out that the indoor channel quality is not only depended on the LoS and NLoS channel state, but the number of times that the signal is blocked by obstacles. The signal attenuation caused by each wall and floor that blocks the propagation of the signal has to be clearly calculated. Thus, the indoor NLoS path loss in dB can be expressed as
\begin{align}\label{PLNLOS}
{L_\text{NLoS}}(d,n) = L_0 + N\log_{10}{d} + L_f(n)     ,
\end{align}
where $N$ is the distance power loss coefficient which depends on the local environments and signal frequency (in office scenario, $N=25.5$ for 2GHz and $N=31$ for 5GHz). $f$ represents the central frequency in MHz and $d$ denotes the separation distance in meter between the AP and IRs. $L_f(n)$ is the floor penetration loss factor in dB and $n$ is the number of obstacles between AP and IRs. As suggested in~\cite{ITUR}, in the office scenario, we have $L_f(n)=15 + 4(n-1)$ for 2GHz and $L_f(n)= 16$ for 5.2GHz. Finally, $L_0$ is the basic transmission loss that qualifies,
\begin{align}\label{L}
L_0 = 20\log_{10}{f} - 28.
\end{align}

It is also pointed out by the ITU-R \cite{ITUR2} that when the receiver is enjoying an indoor LoS channel, the path loss can be expressed as
\begin{align}\label{PLLOS}
{L_\text{LoS}}(d) = 16.9\log_{10}{d} - 27.2 + 20\log_{10}{f} .
\end{align}

Moreover, with a consideration of small scale fading, the channel gain can be expressed as
\begin{align}\label{PL}
g_{u}^ {m}(d) =  L_{u}^ {m}(d) - 10\log_{10}{h_{u}^ {m}},
\end{align}
where $h_{u}^ {m}$ represents random small-scale fading in accordance with the Rayleigh distribution.


\subsection{Signal Model}

Since all IRs are associated with the specified AP $\boldsymbol{m}$, based on NOMA principle, the AP need to transmit a superposition signal that contain information for all IRs $u$
\begin{align}\label{xt}
{x^{\boldsymbol{m}}(t) = \sum\limits_{u = 1}^U {\sqrt {P_{u}^{\boldsymbol{m}}(t)} } x_{u}^{\boldsymbol{m}}(t) },
\end{align}
where $x_{u}^{\boldsymbol{m}}(t)$ represents the transmitted signal for IR $u$ and $P_{u}^{\boldsymbol{m}}(t)$ represents the allocated power for IR $u$. Since we assume that there are several extrinsic interference sources $m \in \mathbb{M}=\{1,2,3...M\}$, the received signals at IR $u$ can be calculated as
\begin{align}\label{yu}
{y_u(t) = g_{u}^ {\boldsymbol{m}}(t)x^{\boldsymbol{m}}(t) + \sum\limits_{m = 1}^M {I^m_u(t)} + \sigma _u^{\boldsymbol{m}}(t) },
\end{align}
where $g_{u}^ {\boldsymbol{m}}(t)$ represents channel gain between AP $\boldsymbol{m}$ and IR$u$, and then $g_{u}^ {\boldsymbol{m}}(t)x^{\boldsymbol{m}}(t)$ is the received superposition signal at IR $u$ from the AP $\boldsymbol{m}$. $I^{m}_u(t)$ denotes the interference from external APs suffered by IR $u$. $\sigma _u^{\boldsymbol{m}}(t)$ represents the additive white Gaussian noise (AWGN).

According to the principle of NOMA, the received superposition signal at IR $u$ comprises its desired signal and the signal for other IRs. Thus, the received superposition signal can be split into
\begin{align}\label{receivedSignal}
g_{u}^ {\boldsymbol{m}}(t)x_u(t) = g_{u}^ {\boldsymbol{m}}(t){\sqrt {P_{u}^{\boldsymbol{m}}(t)} }x_u^{\boldsymbol{m}}(t) + \sum\limits_{s = 1, s\neq u}^U{g_{u}^ {\boldsymbol{m}}(t){\sqrt {P_{s}^{\boldsymbol{m}}(t)} }x_s^{\boldsymbol{m}}(t)},
\end{align}
where $g_{u}^ {\boldsymbol{m}}(t){\sqrt {P_{u}^{\boldsymbol{m}}(t)} }x_u^{\boldsymbol{m}}(t)$ is the desired signal of user $u$ and $\sum\limits_{s = 1, s\neq u}^U{g_{u}^ {\boldsymbol{m}}(t){\sqrt {P_{s}^{\boldsymbol{m}}(t)} }x_s^{\boldsymbol{m}}(t)}$ is the intra-cluster interference received by IR $u$.
Fortunately, a part of the intra-cluster interference can be removed by using successive interference cancellation (SIC) ~\cite{liu2017non}.

The SIC authorizes the receiver with a better channel condition to remove signals for receivers who are in the same NOMA cluster but with weaker channel conditions. To maximize the effectiveness of the intra-cluster interference cancellation, the NOMA receiver has to determine the optimal decoding order.
Specific to this case, since IRs are continuously moving and the indoor channel condition is sensitive with location, there is no IR that always has a strong channel gain. Therefore, a dynamic decoding order has to be considered. The auxiliary term $G_{k}^u(t)$ shown in \eqref{gnku} is interjected as a criterion for determining the decoding order, and $G_{k}^u(t)$ can be regarded as the equivalent channel gain, which can be calculated as
\begin{align}\label{gnku}
G_u^{\boldsymbol{m}}(t) = \frac{g_{u}^{\boldsymbol{m}}(t)}{\sum\limits_{m = 1}^M {I^m(t)} + \sigma _u^{\boldsymbol{m}}(t)}.
\end{align}

Assuming two IRs $j$ and $k$ served by AP ${\boldsymbol{m}}$ are in the same NOMA cluster and their equivalent channel gains can be denoted as $G_{k}^{\boldsymbol{m}}(t), G_{j}^{\boldsymbol{m}}(t)$, respectively. Then SIC can be implement at the IR with a stronger equivalent channel gain to remove the intra cluster interference caused by IR $j$, which can be expressed as
\begin{align}\label{EG}
G_{k}^{\boldsymbol{m}}(t) \ge G_{j}^{\boldsymbol{m}}(t).
\end{align}

Inequation \eqref{EG} is supported by several existing literatures, such as \cite{cui.signal}. Thus, quoting the above theory to this scenario, the decoding order according can be sorted as $G_{\pi (1)}^{\boldsymbol{m}}(t) \le \cdots \le G_{\pi (u)}^{\boldsymbol{m}}(t) \le  \cdots  \le G_{\pi (U)}^{\boldsymbol{m}}(t)$, where $\pi (u)$ denotes the decoding order of IR $u$.

According to the SIC principle, signals for all the IRs 'weaker' than IR  $\pi(u)$ can be decodes and removed at IR $\pi(u)$, after that, the desired signal $x_{\pi(u)}^{\boldsymbol{m}}$ will be decoded. Therefore, the unremoved signals for IR $\pi(u+1)\cdots \pi(U)$ have to be considered as the intra-cluster interference. Thus, the intra-cluster interference of IR $u$ can be calculated as
\begin{align}\label{Iintra}
{\mathcal{I}_{\pi(u)} = \sum\limits_{i=u+1}^{U}g_{\pi (u)}^u(t)\sqrt{P_{\pi (i)}^u(t)}x_{\pi (i)}^u(t)}.
\end{align}

As a consequence of Equation \eqref{yu} \eqref{receivedSignal} and \eqref{Iintra}, the signal-to-interference-plus-noise ratio (SINR) of the IR $u$ can be calculated as \eqref{rnpiku}, where $\mid \mid$ represents the signal power calculation.
\begin{align}\label{rnpiku}
{\gamma _{\pi (k)}^u(t) = \frac{g_{u}^ {\boldsymbol{m}}(t){ P_{u}^{\boldsymbol{m}}(t) }}   { \sum\limits_{i=u+1}^{U}g_{\pi (u)}^u(t){P_{\pi (i)}^u(t)} + \sum\limits_{m = 1}^M {\mid I^m(t)\mid} + {{\mid \sigma _{u}^{\boldsymbol{m}}(t)} ^2} \mid  }}.
\end{align}

As a result, the data rate of IR $u$ can be calculated as
\begin{align}
\mathcal{R}_{u}^{\boldsymbol{m}}(t) = {B}\log 2\left( {1 + \gamma _{\pi (k)}^u(t)} \right),
\end{align}
where $B$ represents the allocated bandwidth for IR $u$. Then, when the data rate does not meet the demand data rate of the IR, this period of time will be determined as outage duration. The total outage duration of IR $u$ can be given by

\begin{align}\label{outage}
 \mathcal T =  \sum\limits_{t=0}^{T_u} \mathcal T_u,
 \mathcal T_u=
\begin{cases}
0& \text{otherwise},\\
1& \mathcal{R}_{u}^{\boldsymbol{m}}(t)< \mathcal{R}_d ,
\end{cases}
\end{align}
where $\mathcal T$ denotes the outage duration $T_u$ represents the mission duration and $\mathcal{R}_d$ represents the demand data rate of the IR.
\subsection{Mission Quality Indicator}

In order to accurately evaluate the performance of IRs in completing tasks, we propose a key indicator, namely mission quality indicator (MQI). The MQI includes two factors, the time cost on the path and the time duration that IRs cannot guarantee a qualified connection with the AP. Therefore, our proposed MQI considers both time costs of the mission and IR outage time duration, which can be express as

\begin{align}\label{MQI}
\text{MQI} = T_\text{total}-\sum_{u=1}^{U}(T_u+ {\lambda \mathcal T_u}),
\end{align}
where $T_\text{total}$ denotes the maximum mission time and $T_u$ denotes the time IR $u$ spent on the way. $\lambda$ represents the outage penalty coefficient, which can be selected according to the reliability requirement of the mission.


\subsection{Problem Formulation}

With the assistance of the indoor radio map, we aim to maximize the MQI of IRs by optimizing their path $\mathbb{D} = \{D_u(1),D_u(2),...D_u(t)...D_u^d\}$ and the corresponding power allocation policy $\mathbb{P}= \{P^{\boldsymbol{m}}_u(1),P^{\boldsymbol{m}}_u(2),...P^{\boldsymbol{m}}_u(t)\}$, where $D_u(t)=[x_u(t),y_u(t),z_u(t)]$ represents the position of IR $u$ at time $t$ and $D_u^d$ represents the destination of IR $u$ . The optimization problem is formulated as \eqref{OPP}, where MQI is depended on the weighted sum mission time and outage duration.
\begin{subequations}
\begin{align}\label{OPP}
\max_{\mathbb{D,P}} \quad & T_\text{total}-\sum_{u=1}^{U}(T_u + \lambda \mathcal T_u), \\
\textrm{s.t.}
& z_u(t) = 0, \forall u \in \mathbb{U},\forall t \label{OPPB}\\
& \mathbb{D}_u \geqq 1, \forall u \in \mathbb{U}, \label{OPPC}\\
&{x_{\min }} \le x_u(t) \le {x_{\max }},\forall u,\forall t,\notag\\
&{y_{\min }} \le y_u(t) \le {y_{\max }},\forall u,\forall t \label{OPPD},\\
&\sum\limits_{u \in {\mathbb U}} { {P^{\boldsymbol{m}}_u(t) \le {P}^{\boldsymbol{m}}(t)} }, \forall t,\forall u,\label{OPPE}\\
&G_{\pi (k)}^u \ge G_{\pi (j)}^u,k > j,\forall (k,j),\forall t,\forall u,\label{OPPF}\\
& \mathcal{R}_{u}^{\boldsymbol{m}}(t)> \mathcal{R}_d , t \notin \mathcal T_u.\label{OPPG}
\end{align}
\end{subequations}

There are several constraints listed in the optimization problem. Constraint \eqref{OPPB} illustrates that the IRs have to move on the floor and their path cannot go through anywhere unreachable or blocked. Constraint \eqref{OPPC} requires that there is at least one valid path to each mission point. Constraint \eqref{OPPD} ensures that IRs are always kept in the selected room during the mission duration. Constraint \eqref{OPPE} is a power constraint that ensures the sum transmitting power for each IR does not exceed the maximum transmit power of the AP. Constraint \eqref{OPPF} represents that the optimal decoding order suggested in Equation \eqref{EG} is safeguarded during the mission. Finally, constraint \eqref{OPPG} qualifies the minimum data rate requirement.

\section{Indoor Radio Map}

In general, the radio map indicates the power distribution of the transmitted signal in a given area ~\cite{bi2019engineering}. \footnote{In engineering, the radio map of a specific area is often measured and reconstructed by interpolation ~\cite{zou2017winips}.} The power distribution map is constructed according to the path loss and the fading expectation, since the fading has high randomness and is non-trivial to be estimated accurately. Therefore, the radio map can represent the statistical channel quality of each position, but it cannot be considered as completely accurate environmental knowledge at any specific time.
However, the power distribution of the interferences needs to be considered as well. Thus, an SINR map can be derived according to the power distribution maps of AP $\boldsymbol{m}$ and interferer $m$. The SINR map is employed to provide knowledge to figure out the power allocation for the NOMA cluster and guide the path for IRs.

\subsection{Radio Map Based Training Process}

The introduction of the radio map is to provide an environment for the agent training. Prior to starting missions of IRs, the radio map of the given mission area has to be constructed and input into the agent. After that, the agent can be virtually trained that the agent imagines acting in the mission area, estimates the data rate based on the radio map and calculates rewards. With this training paradigm, IRs do not need to perform any actual action in reality during the training process, which is more economical and energy-efficient compare to training in reality. It may worth noting that the activities of people in the mission area do not have a significant impact on indoor signal propagation. Therefore, the pre-measured radio map is always effective unless the interior furnishings are changed. Moreover, the radio map training with the expected fading model is more conducive to agent training, since the agent can know the expected fading scale without being disturbed by the uncertainty. Therefore, due to the superiorities mentioned above, we invoke the radio map to train the agent.

\subsection{3D Interior Layout Map}

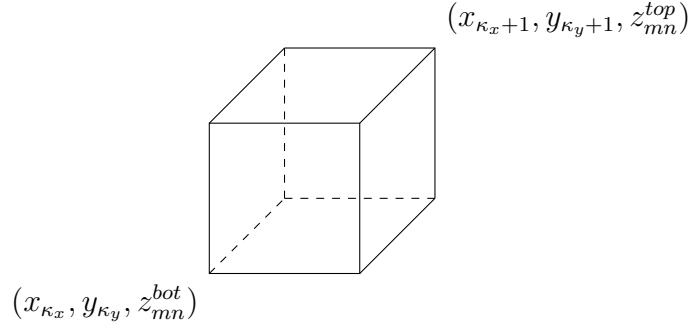
\begin{figure}[t!]	
\centering	
\begin{tikzpicture}
     \coordinate (A) at (0,0) node[below left] at (A) {$(x_{\kappa_x},y_{\kappa_y},z_{mn}^{bot})$};
     \coordinate (B) at (2,0) node[below right] at (B) {};
     \coordinate (C) at (2,2) node[right] at (C) {};
     \coordinate (D) at (0,2) node[left] at (D) {};
	
     \coordinate (E) at (1,1) node[left] at (E) {};
     \coordinate (F) at (3,1) node[right] at (F) {};
     \coordinate (G) at (3,3) node[above right] at (G) {$(x_{\kappa_x+1},y_{\kappa_y+1},z_{mn}^{top})$};
     \coordinate (H) at (1,3) node[above left] at (H) {};
	
			
     \draw (A) rectangle (C);
     \draw[dashed] (E)--(F);
     \draw (F)--(G)--(H);
     \draw[dashed] (H)--(E);
     \draw[dashed] (A)--(E);
     \draw (B)--(F);
     \draw (C)--(G);
     \draw (D)--(H);
	
	
   \end{tikzpicture}	
   \caption{Cuboid $\mathcal{L}_{mnz}$ in the interior layout map.}
	\label{Figbox}
 \end{figure}

\begin{figure*}[t!] 
\centering 
\includegraphics[width=0.8\textwidth]{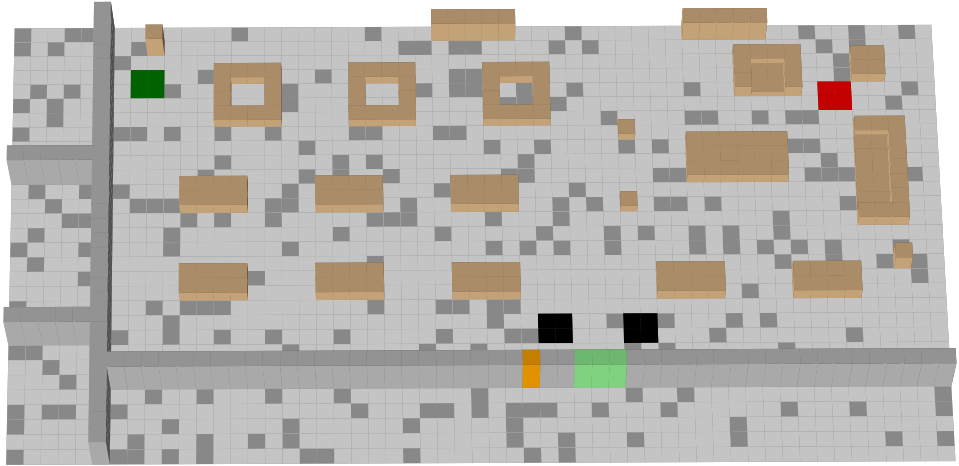} 
\caption{3D interior layout map.} 
\label{Fig.3DS} 
\end{figure*}

In order to generate a radio map, we first need to create a deterministic interior layout, which includes models of walls, furniture and other objects that may affect the propagation of radio waves. We propose a Lego modeling method for building indoor models with virtual bricks. A 3D interior model created by this method is displayed in Fig~\ref{Fig.3DS}, which is a digital description of the room shown in Figure~\ref{Fig.main2}. A number of bricks like Lego toys are engaged to outline the interior walls and furniture. Firstly, the entire room space has to be discretized into a grid map according to a selected resolution $\kappa_x,\kappa_y$. We denote the grid square with vertex coordinates $(x_{\kappa_x},y_{\kappa_y}),(x_{\kappa_x+1},y_{\kappa_y + 1})$ as $\mathcal{L}_{\kappa_x\kappa_y}$, where $1$ is the unit side length of a square and $\kappa_x,\kappa_y$ denotes the row and column order of the squares. Then we delineate the shape of the object that may deleterious the LoS path as a cuboid, which can be denoted as $\mathcal{L}_{mnz}$ with the vertex coordinates $(x_{\kappa_x},y_{\kappa_y},z_{mn}^{bot}),(x_{\kappa_x+1},y_{\kappa_y+1},z_{mn}^{top})$ as shown in Fig. \ref{Figbox}, where $z_{mn}^{bot} $ and $z_{mn}^{top}$ can be considered as the hight of bottom and top of the cuboid. Perform the above modeling for objects in each grid, the indoor surrounding map can be expressed as two matrixes which contain the profile of the interior layout

\begin{equation}
\mathcal{L}_\text{bottom}= \left( \begin{array}{ccc}
z_{11}^{bot} & \cdots & z_{1\kappa_y}^{bot}\\
\vdots& \ddots &\vdots \\
z_{\kappa_x}^{bot} & \cdots & z_{\kappa_x\kappa_y}^{top}
\end{array} \right),\qquad
\mathcal{L}_\text{top}= \left( \begin{array}{ccc}
z_{11}^{top} & \cdots & z_{1\kappa_y}^{top}\\
\vdots& \ddots &\vdots \\
z_{\kappa_x1}^{top} & \cdots & z_{\kappa_x\kappa_y}^{top}
\end{array} \right).
\end{equation}

Please note that in this case, we ignored hanging objects such as chandeliers and air conditioners, since these objects with higher height are less likely to block signals of IRs standing on the ground, but it is worth noting that the proposed Lego modeling method is capable to describe floating objects if necessary. Moreover, in this method, a higher-resolution model with smaller bricks can depict the outline of the objects more accurately, but it greatly increases the computational complexity as well.

\subsection{Detection of Occlusion}

In order to construct the power distribution map of the AP, it is necessary to primarily determine whether there is an LoS propagation path in each position of the room. According to the generated 3D layout model, the channel conditions at each grid in the room can be determined. Assuming that the current position of IR is $D_{u} = \{x_u, y_u, z_u\}$ and the position of the AP is $D_{\boldsymbol{m}} = \{x_{\boldsymbol{m}}, y_{\boldsymbol{m}}, z_{\boldsymbol{m}}\}$, the vector $\overrightarrow{D_{\boldsymbol{m}} D_u}$ can be used to represent the direct propagation path of the signal. It is difficult to directly determine whether the vector and irregular objects such as the sofa have any intersection. Instead, with the aid of the layout profile matrix, we are able to calculate whether the vector $\overrightarrow{D_{\boldsymbol{m}}D_u}$ has an intersection with the cuboid $\mathcal{L}_{mnz}, \forall m, \forall n$. As a typical ray-box intersection detection problem, this can be solved by the axis-aligned bounding box (AABB) algorithm \cite{majercik2018ray}, which is widely acknowledged in the computer science field.
By traversing all the cuboids, we can determine whether the measured path is LoS or NLoS, and then we can further tag each square as with LoS or not.
After the LoS state of each grid are collected, it can be used to build the power distribution map.

\subsection{Power distribution Map}

The power distribution map directly reflects the power intensity of the signal transmitted by the AP at different receiving position. In order to calculate and store the power distribution map in the digital system, we divide the 2D map into equally spaced grids as what we did in the surrounding map. Sampling points is set at the center of each grid square and the coordinate noted as $x,y$ inside the square room, where $x_\text{min} <x< x_\text{max}, y_\text{min} <y< y_\text{max}$.
The first step of radio map construction is determining if there is LoS or how many obstacles are there between the transmitter and the receiver according to the interior layout map and the occlusion detection. After that,
based on the aforementioned indoor propagation model, the path loss for each grid can be calculated as Equation \eqref{Radio.map}.

\begin{align}\label{Radio.map}
g_{x,y}^{\boldsymbol{m}} =
\begin{cases}
16.9\log_{10}{d^{\boldsymbol{m}}_{x,y}} - 27.2 + 20\log_{10}{f}-\mathbf{E}(h^{\boldsymbol{m}}_{x,y}), & {\kern 4pt} \text{if LoS},\\
20\log_{10}{f} - 28 + + N\log_{10}{d^{\boldsymbol{m}}_{x,y}} + L_f(n)-\mathbf{E}(h^{\boldsymbol{m}}_{x,y}), & {\kern 4pt} \text{if NLoS},\\
0 , & {\kern 4pt} \text{otherwise},
\end{cases}
\end{align}

\begin{align}\label{d}
{d_{x,y} = \sqrt {{h}_{\boldsymbol{m}}^2 + {{\left[ {{x} - {x_{\boldsymbol{m}}}} \right]}^2} + {{\left[ {{y} - {y_{\boldsymbol{m}}}} \right]}^2}} },
\end{align}
where $\mathbf{E}(h^{\boldsymbol{m}}_{x,y})$ represents the excepted fading scale, $d_{x,y}$ represents the distance between measuring point and AP $\boldsymbol{m}$, the $h_{\boldsymbol{m}}$ represents the altitude intercept of AP $\boldsymbol{m}$ and the IR antenna.

\begin{figure} \centering
\subfigure[Power distribution map] {
 \label{Fig.powermap}
\includegraphics[width=0.47\columnwidth]{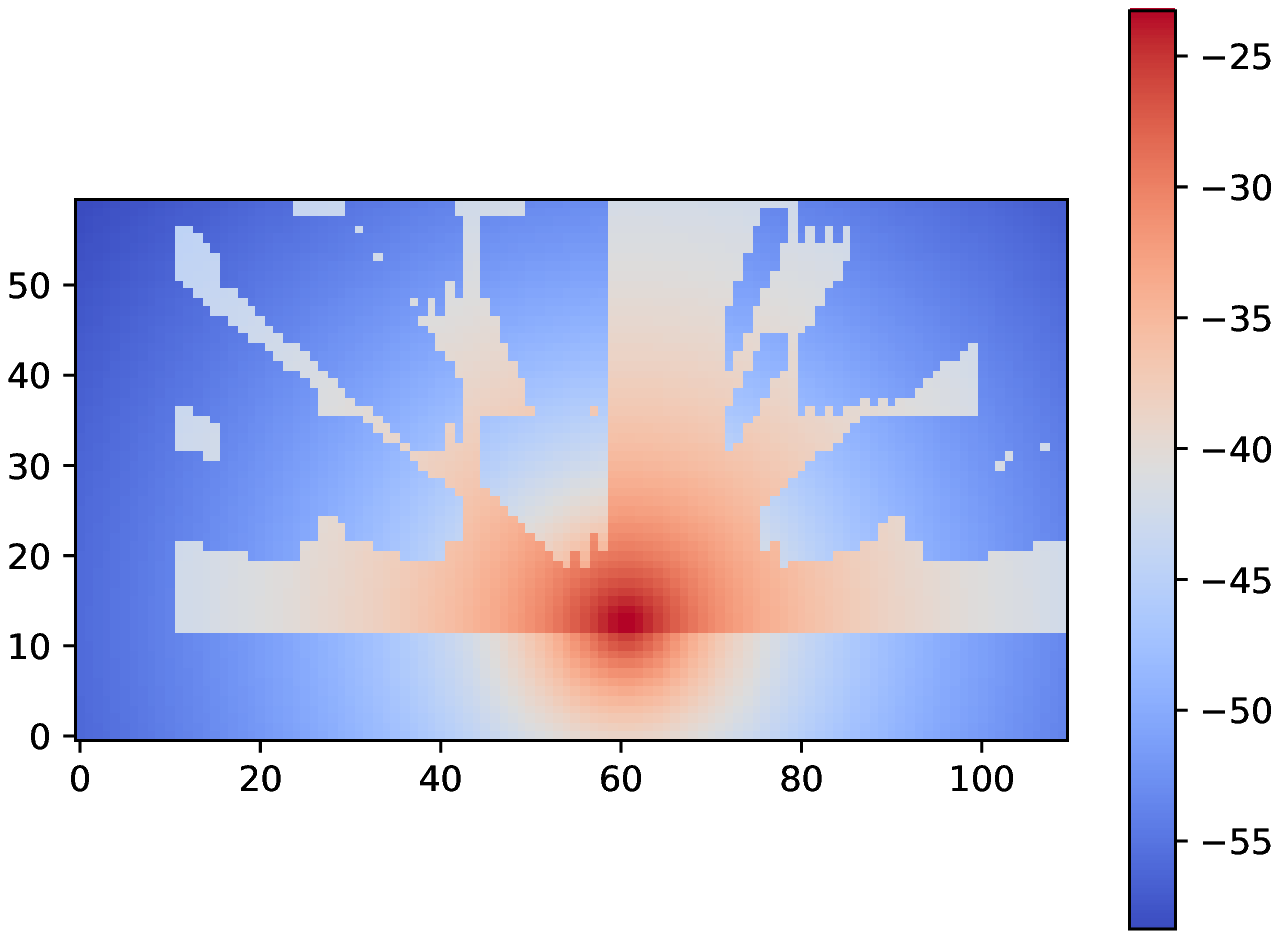}
}
\subfigure[SINR map] {
\label{Fig.SINRMAP}
\includegraphics[width=0.47\columnwidth]{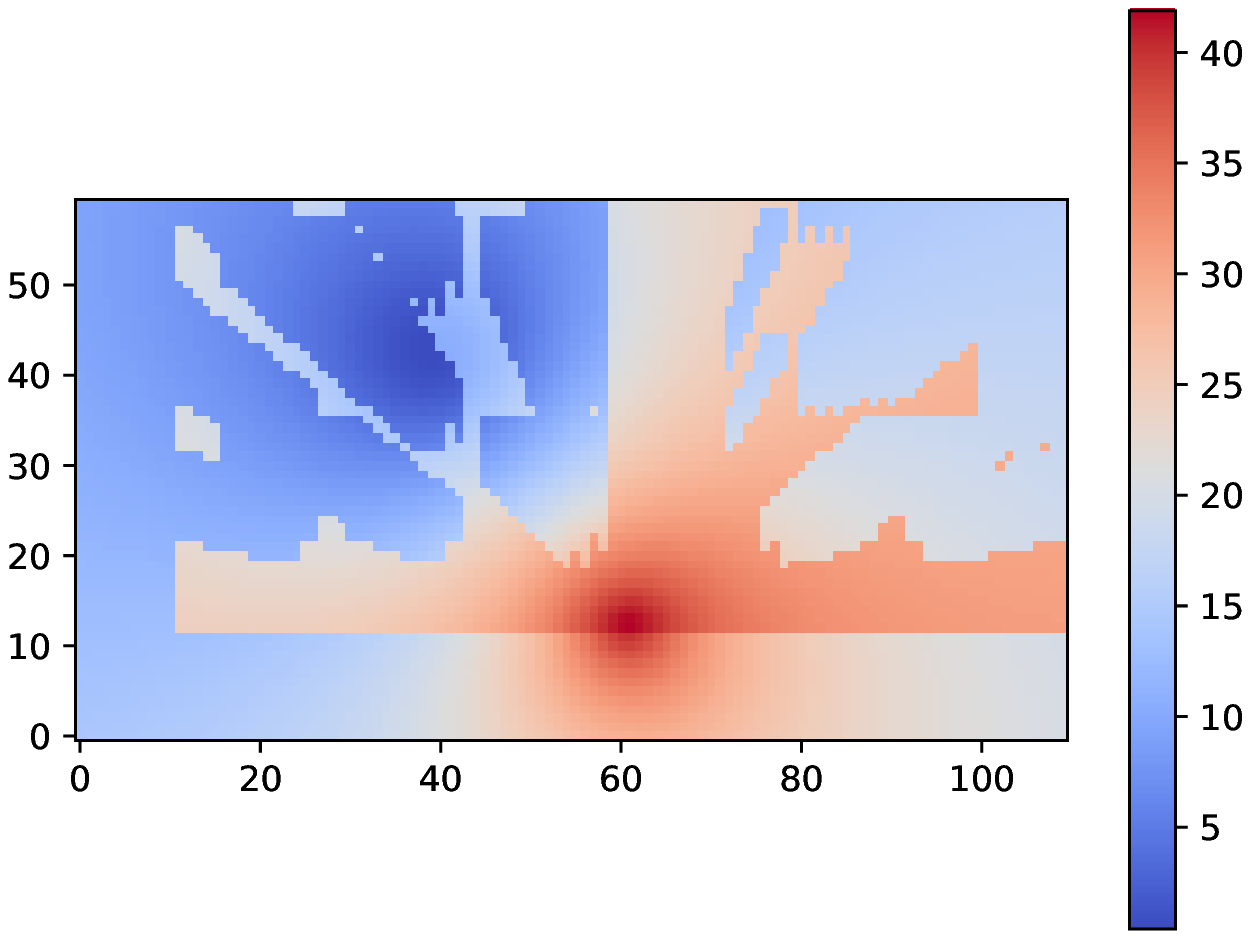}
}

\caption{Indoor radio maps of the mission area}
\label{fig}
\end{figure}

Therefore, the power distribution map of the AP can be established with Equation \eqref{Radio.map} and a given transmitting power. Since the IR moves at a fixed altitude, the 2D radio map can be expressed as in Fig. \ref{Fig.powermap}. The power distribution map has a significant regular pattern and it can be observed that the strength of the received signal is highly related to the distance and whether there is LoS.

\subsection{SINR Map}

The power distribution map is not enough to describe the channel quality of each receiving position, since external interferences have not been considered. Thus, the SINR map containing interference information has to be derived for agent training. With the power distribution maps of AP $\boldsymbol{m}$ and interferer $m$, the SINR at each grid can be calculated as

\begin{align}\label{SINR.map}
SINR_{x,y}^{\boldsymbol{m}} =
\begin{cases}
\frac{P^{\boldsymbol{m}} \sum\limits_{m = 1}^M{g_{x,y}^{m}}/P^m}{g_{x,y}^{\boldsymbol{m}}} & {\kern 4pt} x_\text{min}<x<x_\text{max} , y_\text{min}<y<y_\text{max},\\
0 , & {\kern 4pt} \text{if otherwise}.
\end{cases}
\end{align}

Then, according to \eqref{SINR.map}, the SINR map shown as Fig. \ref{Fig.SINRMAP} can be constructed. It can be observed that the SINR map shows more complicated changes compared to the power distribution map. The interference source located on the upper floor formed an SINR abyss in the (40, 40) area and another outdoor interference source also significantly lower the transmission quality on the right side of the room.

\section{The Proposed Algorithm for Path Designing and Resource Allocation}

\subsection{Motivation}

We propose the DT-DPG algorithm since the existing RL algorithms expose two vulnerabilities in the optimization problem \ref{OPP}. Prior to the main task, planning actions and power allocation, the precondition of the motion planning is that the agent needs to find out the location of the destination point. Alternatively, we can enforce the agent to the destination through a fairly long random training process and high rewards to reach the destination. However, in a complex indoor environment, the amount of training steps required for this scheme is almost unacceptable.

Furthermore, since the agent will be punished when it consumes time or cannot meet the communication quality, this suggests that the agent of the conventional DDPG algorithm is expected to continue to receive negative rewards when it takes action. Therefore, the agent has a noticeable probability to succumb to the negative rewards and refuse to explore the destination, especially when the destination is in the SINR basin. In fact, this is a common problem of RL algorithms that when confronting a complex environment or problem, the agent is likely to fall into the local optimum and do not explore the global optimal solution \cite{matheron2019problem} \cite{li2019robust}. In order to solve the above-mentioned problems, we proposed the DT-DPG algorithm.

\subsection{DT-DPG algorithm for motion planing and power allocation}
The proposed DT-DPG algorithm is a hybrid algorithm of the conventional DDPG algorithm and the transfer learning, in which we also made improvements on the Ornstein Uhlenbeck (OU) action noise and neural network structure of the original DDPG algorithm described in \cite{lillicrap2015continuous}. We assume that there is a central controller attached with the AP playing a role as an agent, responsible for the power allocation and determining all IRs' motions.  Following a classic reinforcement learning model for the Markov process, the agent observes the current state $S_t$ and implements action $A_t$. As a consequence of the action, the state will move to $S_{t+1}$, and the agent will get reward $R_t$ as the feedback. We hire two pairs of neural networks, including actor network $\mu$, critic network $\theta$ and their corresponding target network $\mu',\theta'$. The participation of the target network is to prevent the neural network from divergent during training.
Then, the action can be calculated by the action function and the OU noise \cite{colas2018gep} as
\begin{align}\label{Action}
A_t = \mu(S_t|\theta_t^\mu)+N(0,\xi),
\end{align}
where the $N(0,\xi)\sim \xi(\mathcal{X},\mathcal{E})$ represent the OU noise, and $\mathcal{X} \in \mathbb{X}$ and $\mathcal{E} \in \mathbb{E}$ denote the training task and the episode number, respectively. In this paper, we adopt linearly diminishing OU noise instead of a fixed OU noise scale. The initial scale $\xi(\mathcal{X},0)$ of OU noise is determined according to the requirements of the training task $\mathcal{X}$. This scheme allows the agent to better explore the environment and maintain stable convergence in the later stages of training. Generally, when $\mathcal{X}=0$, $\xi$ should take a larger value, which encourages the agent to take actions more randomly to execute sufficient exploration.

After the determined action is executed in the environment, the agent saves the experience as $(S_t,A_t,R_t,S_{t+1})$ to the memory bank to be sampled to train the neural network.
For a single action, the actor network are updated by using
\begin{align}\label{Actor}
\nabla_{\theta^\mu}J = \nabla_A \mathcal{Q}(S,A|\theta^\mathcal{Q})\nabla_\theta^\mu \mu(S|\theta^\mu).
\end{align}

Then, the critic network is updated by minimizing the loss
\begin{align}\label{Loss}
Loss_t = (y - Q(S,A,w))^2,
\end{align}
where
\begin{align}\label{Y}
y_t = R_t + \rho  \mathcal{Q}'(S_{t+1},\mu'(S_{t+1}|\theta^{\mu'})|\theta^{\mathcal{Q}'}).
\end{align}

\begin{figure}[htbp]
\centering
\includegraphics[width=1.0\columnwidth]{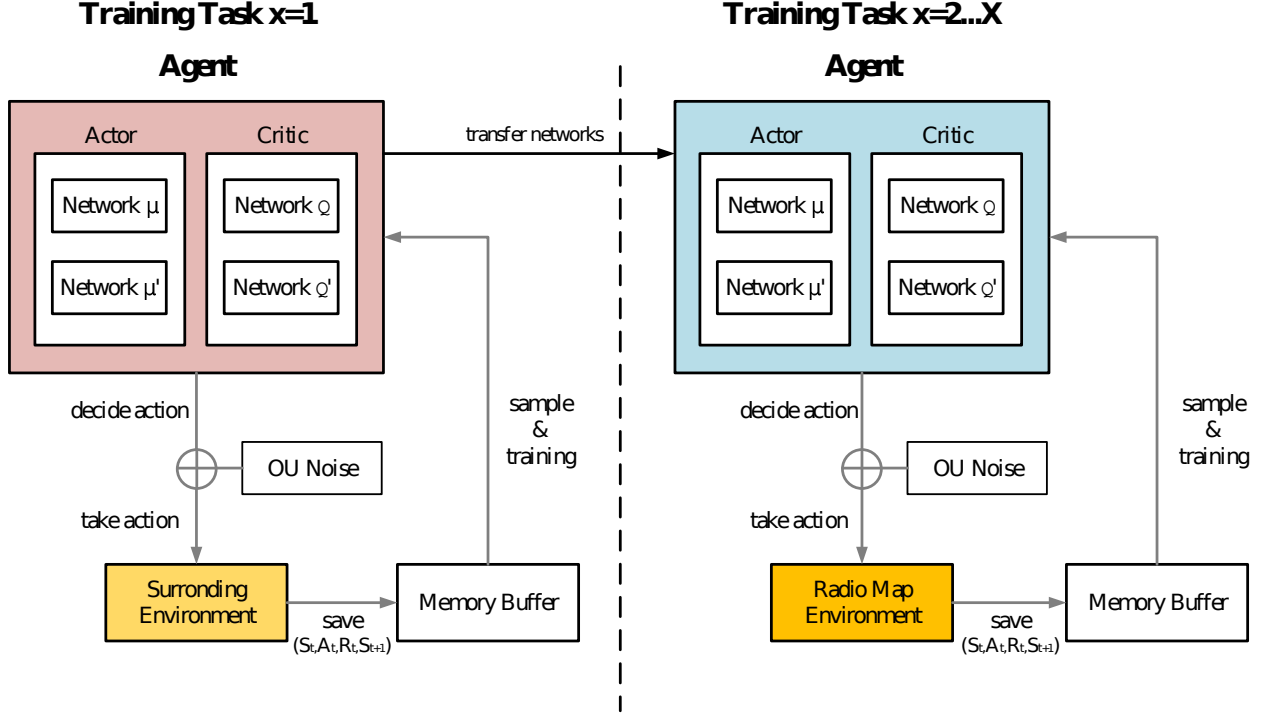}
\caption{Flow diagram of the proposed DT-DPG algorithm}
\end{figure}

\begin{algorithm}
\caption{DTDPG algorithm for the motion planing and power allocation}
\label{DTDPG}
\begin{algorithmic}[1]

        \FOR{each training task $\mathcal{X}$}
        \STATE Initialize the environment $E_\mathcal{X}$, reward function $R_\mathcal{X}$ and input the radio map
        \IF {$x$ = 1}
            \STATE Initialize the actor network $\mathcal{Q}(S,\theta^\mathcal{Q})$, critic network $\mu(S,\theta^\mu)$, target actor network$\mathcal{Q}'(S,\theta^\mathcal{Q'})$, target critic network $\mu'(S,\theta^{\mu'})$ with random parameter $\theta^\mathcal{Q}$, $\theta^\mu$, $\theta^\mathcal{Q'}$, $\theta^{\mu'}$
        \ELSE
            \STATE Initialize the actor network $\mathcal{Q}(S,\theta^\mathcal{Q})$, critic network $\mu(S,\theta^\mu)$, target actor network$\mathcal{Q}'(S,\theta^\mathcal{Q'})$, target critic network $\mu'(S,\theta^{\mu'})$ with the trained networks $\theta^\mathcal{Q}_{(\mathcal{X}-1)}$, $\theta^\mu_{(\mathcal{X}-1)}$, $\theta^\mathcal{Q'}_{(\mathcal{X}-1)}$, $\theta^{\mu'}_{(\mathcal{X}-1)}$
        \ENDIF
        \FOR{each episode}
        \STATE Reset initial positions and destinations for IRs
        \STATE Update the standard deviation $N$ of the action noise
        \FOR{each step in $ t_0 \leq t \leq t_\text{max}$}
        \STATE Observe $S_t$ according to the radio map
        \STATE Choose $A$ according to action policy and $Q(S,\theta^\mathcal{Q})$
        \STATE IRs take action $A$, observe $R$ and $S'$
        \STATE Store $e = (S_t,A,R,S_{t+1})$
        \STATE Random sample a batch of $e$ from memory buffer

        \STATE Calculate target according to Equation \eqref{Y}
        \STATE Train critic network $\mu(S,\theta^\mu)$ with a gradient descent step Equation \eqref{Loss}
        \STATE Train actor network $\mathcal{Q}(S,\theta^\mathcal{Q})$ with Equation \eqref{Actor}
        \STATE Update the target networks$\theta^\mathcal{Q'}\leftarrow (1-\tau)\theta^\mathcal{Q'} + \tau\theta^\mathcal{Q}$, $\theta^\mathcal{\mu'}\leftarrow (1-\tau)\theta^\mathcal{\mu'} + \tau\theta^\mathcal{\mu}$
        \STATE $S_t \leftarrow S_{t+1}$
        \ENDFOR
        \STATE Save the networks $\theta^\mathcal{Q}$, $\theta^\mu$, $\theta^\mathcal{Q'}$, $\theta^{\mu'}$ as $\theta^\mathcal{Q}_\mathcal{X}$, $\theta^\mu_\mathcal{X}$, $\theta^\mathcal{Q'}_\mathcal{X}$, $\theta^{\mu'}_\mathcal{X}$
		\ENDFOR
        \ENDFOR

\end{algorithmic}
\end{algorithm}

\subsection{Neural Network Structure for DT-DPG algorithm}

\begin{figure}[htbp]
\centering
\includegraphics[width=0.8\columnwidth]{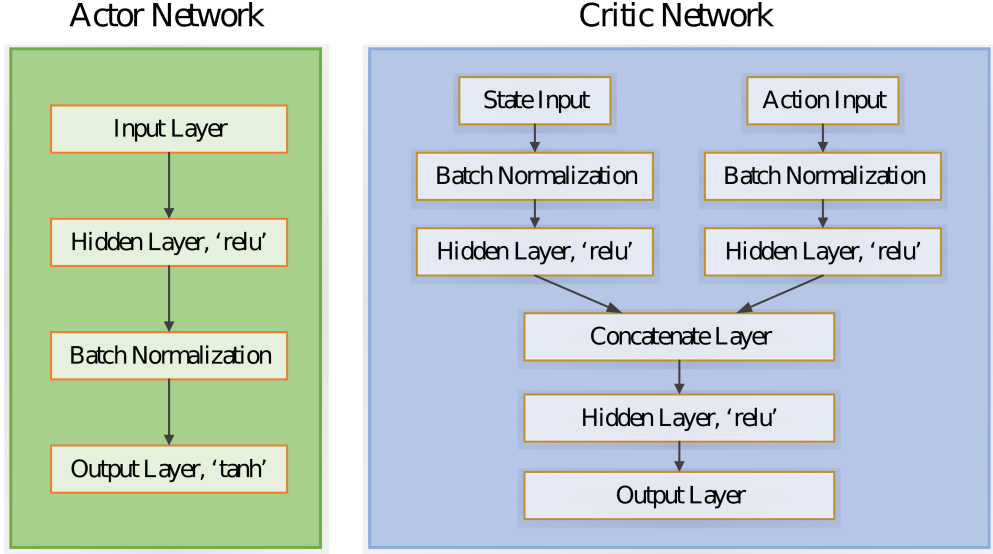}
\caption{Neural network structure of the proposed DT-DPG algorithm} \label{Fig.NN}
\end{figure}

We adopt structures with multiple hidden layers and batch normalization (BN) layers as the actor network and critic network, which is displayed in Fig \ref{Fig.NN}. The activation function 'relu' is selected for all activation layers. In order to ensure normalized output actions, an output layer with activation function 'tanh' is employed in the actor network.
In the critic network, both the state and the action are input, thus, the scales of the input values are likely to be significantly different. Therefore, the input data have to path through a BN layer prior to other hidden layers. Then, the two inputs are concatenated in a concatenate layer. After passing through activation layers, an estimated value of the action will be output by the critic network.

Lillicrap et al. added BN layers before each layer in their original DDPG algorithm \cite{lillicrap2015continuous}. However, we want to highlight the specialty of adding the BN layer before the output layer with the activation function 'tanh'. The BN layer in the actor network before the output layer ensures that the input of the 'tanh' function is in the valid range, while the purpose of the BN layer in the critic network is to normalize two types of input data with different scales. Please refer to Section \uppercase\expandafter{\romannumeral5} for specific neural network and parameter settings.

\begin{remark}
The BN layer can prevent the input value of the 'tanh' layer from exceeding the valid range. Once the BN layer before the output layer is absent, the agent is likely to output extreme actions catastrophically.
\end{remark}

\subsection{State and Action Space for DT-DPG algorithm}

To solve the formulated problem, we design the state space and action space as follows. Thanks to the actor-critic scheme, the agent armed with the DT-DPG model is capable to adopt continuous state space and action domain, which allows the IR to adopt a more sleek move and precise power distribution.

\subsubsection{State Space}

The agent constructs state $S_t$ by observing the current location $D_{u}(t)$ of each IR. Meanwhile, the channel gain $g^{{\boldsymbol{m}}}_u(t)$ described in Equation \eqref{PL} is another component of $S_t$, which is considered as the basis for the power allocation. Therefore, the input state can be express as
\begin{align}\label{S}
S_t = \{D_{u}(t),g^{{\boldsymbol{m}}}_u(t)\}, u \in \mathbb{U}.
\end{align}

\subsubsection{Action Space}

Corresponding to the input state $S_t$, the action space contains two subsets, the IRs' motion and power allocation policies at time slot $t$.
Then, the agent has the following action domain:
\begin{align}\label{S}
 A_t = \{\Delta D_{u}(t),P_{\boldsymbol{m}}^u(t)\}, u \in \mathbb{U},
\end{align}
where $\Delta D_{u}(t) = \{\Delta x_{u}(t),\Delta y_{u}(t),0\}$ since we assume that IRs always move on the ground.
\begin{itemize}
\item Motion: The agent is charged to choose motion for each IR. The amplitude of $\Delta D_{u}(t)$ can be arbitrarily selected in $[0-V_\text{max}]$. When the IR plans to travel toward furniture or walls, the component of action perpendicular to the contact surface will be cancelled to avoid collision.
\item Power allocation: The agent outputs power $0 < P^{\boldsymbol{m}}_{\pi(u)}(t)< P^{\boldsymbol{m}}_{\text{max}\pi(u)}$ as the power allocated to each IR. Considering the characteristics of the NOMA, the maximum power of each IRs can be set differently, but $\sum\limits_{u \in {\mathbb U}} { P^{\boldsymbol{m}}_{\text{max}\pi(u)} \le {P}^{\boldsymbol{m}}_\text{max} }, \forall u,$ have to be guaranteed.
\end{itemize}

\subsection{Reward Function for DT-DPG algorithm}

\subsubsection{Destination training(The First Training Phase)}

In order for IRs to quickly find out their destination, we artificially add an inducing reward for this stage of training, where the reward is inversely proportional to the distance between the IR and the corresponding destination.
Therefore, we set several rewards $R_c(t) = -1 $, $R_i(t) = 1 - (d / dmax)$, $R_d = 10$, $R_s = 200$ where $R_t$ is the time cost to spur on the IRs. $R_i(t)$ is the inducing reward to guide the IRs to find out their distention, where $d$ represents the current distance between the IR and destination and $d_\text{max}$ is the maximum possible distance between the IR and destination. In order to make the IRs approach the destination instead of wandering, the step reward design has to follow $R_c + R_i < 0$.

When a single IR arrives at the destination, it will get the reward $R_d$ and stop moving. While all IRs reach their destinations, the agent will obtain reward $R_s$ for the successful mission and end the episode. Summarily, the reward function in destination training can be defined as
\begin{equation}\label{R}
R_1=
\begin{cases}
R^u_c(t) + R^u_i(t), &  \text{out of destination},\\
R^u_c(t) + R^u_i(t) + R^u_d, &  \text{arriving destination },\\
R^u_c(t) + R^u_i(t) + R^u_d + R_s,  & \text{all IRs arrived destinations}.
\end{cases}
\end{equation}

\subsubsection{MQI Training (The Second Training Phase)}

It is worth noting that the motion and power allocation are jointly optimized in this training stage, which cannot be regarded as power optimization based on the optimal path since the path is still changeable in this training phase. In fact, exploration has to be restarted at the beginning of this training phase, and the agent will attempt motions under the new reward regime. In this stage of training, the reward design has to be exactly the same as the optimization goals. Thus, we replace inducement reward $R_i(t)$ by the QoS reward $R^u_\text{QoS}(t)$. Whenever an IR violates QoS requirements at a single time slot, the agent will receive a negative reward $R^u_\text{QoS}(t) = -\lambda$, which can be easily derive from \eqref{MQI} and $R_c(t)$. Therefore, the reward function of the MQI training can be summarized as

\begin{equation}\label{R}
R_2=
\begin{cases}
R^u_c(t) + R^u_\text{QoS}(t), &  \text{out of destination},\\
R^u_c(t) + R^u_\text{QoS}(t) + R^u_d, &  \text{arriving destination },\\
R^u_c(t) + R^u_\text{QoS}(t) + R^u_d + R_s,  & \text{all IRs arrived destinations}.
\end{cases}
\end{equation}

\begin{remark}
Falling into a local optimum is a significant defect of the DDPG algorithm, especially when facing a complex reward mechanism \cite{matheron2019problem} \cite{li2019robust}. In our model, the conventional DDPG algorithm is likely to get sequacious in the negative rewards and never reach the destination. The proposed DT-DPG algorithm effectively solves this problem with the aid of transfer learning. Since the optimization object is disassembled in each training phase, the agent has a clear objective and is capable to solve it.
\end{remark}


\subsection{Convergence and Complexity Analysis}
\subsubsection{Convergence}

In contrast to iterative algorithms, the neural network function approximators introduced in policy based algorithms make convergence unguaranteed \cite{lillicrap2015continuous}. Therefore, we cannot give the proof of convergence, but the numerical results are exhibited to prove that the proposed algorithm is capable to converge under appropriate conditions. The main conditions for the neural network to have a reliable convergence and stability is that the learning rate and the update rate $\tau$ of the target network have to be in an appropriate range. In addition, as pointed out in \textbf{Remark \ref{lr}}, a huge learning rate difference between the actor network and the critic network is not conducive to the convergence of the algorithm.


\begin{remark} \label{lr}
A number of experimental results suggest that if there is a significant gap between the actor learning rate and the critic learning rate, the convergence of the algorithm will be challenged, which may make the reward oscillate or even fail to converge.
\end{remark}

\subsubsection{Complexity Analysis}

This section first describes the computational complexity required for a single decision of the neural network, and then introduce the computational complexity of the proposed algorithm from an overall perspective.

\begin{itemize}
\item The complexity of a single decision: In the proposed algorithm, the neural network consists of normalization layers, 'relu' layers, and  'tanh' layers. Assume that the network $\theta$ has $\vartheta$ fully connected layers, $\theta_\text{n}$ normalized nodes, $\theta_\text{r}$ 'relu' nodes and $\theta_\text{t}$ 'tanh' nodes, than the total floating point operation times can be express as
    \begin{align}
    \mathcal{C_\theta}= 5 \cdot \theta_\text{n} + \theta_\text{r} + 6 \cdot \theta_\text{t} + \sum \limits_{\vartheta_i=0}^{\vartheta}\|\vartheta_i\|\cdot \|\vartheta_{i+1}\|,
    \end{align}
    where $\sum \limits_{\vartheta_i=0}^{\vartheta}\|\vartheta_i\|\cdot \|\vartheta_{i+1}\|$ represents the complexity of adding bias \cite{qiu2019deep} and $\|\vartheta_i\|$ represents the node number of layer $i$. Based on the above calculation, we denote the computational complexity of a single decision of the actor network as $\mathcal{C_Q}$ and the critic network as $\mathcal{C}_\mu$.

\item The complexity of DT-DPG algorithm: In the training process, each time slot the actor network needs to select an action, and the complexity is $\mathcal{O}(\mathcal{C_Q})$. Correspondingly, the complexity caused by the training is $\mathcal{O}(\|e\| \cdot (\mathcal{C}_\mu + \mathcal{Q}_\text{r,t} + \sum \limits_{\mathcal{Q}_i=0}^{\mathcal{Q}}\|\mathcal{Q}_i\|\cdot \|\mathcal{Q}_{i+1}\| + \mu_\text{r,t} + \sum \limits_{\mu_i=0}^{\mu}\|\mu_i\|\cdot \|\mu_{i+1}\|))$. The complexity of updating target network is $\mathcal{O}(\mathcal{Q}_\text{r,t}+\mu_\text{r,t})$. Hence, The computational complexity of a single time slot is approximately $\mathcal{O}((1+\|e\|) \cdot \mathcal{C_Q} + 2 \cdot \|e\| \cdot \mathcal{C}_\mu)$. Thus, the total amount of calculation required for training is $\mathcal{O}(n \cdot t \cdot ((1+\|e\|) \cdot \mathcal{C_Q} + 2 \cdot \|e\| \cdot \mathcal{C}_\mu))$, where $n$ is episodes number and $t$ is the time spent in each episode. The author hopes to stress that since the DT-DPG algorithm can quickly find the destination at the early stage of training, which means the agent spends less average time $\bar{t}$ per episode since the episode will be ended when all IRs reach their destinations. As a consequence, the DT-DPG algorithm requires less training time than the DDPG algorithm for the given problem while the complexity of transferring the NN is negligible.

\end{itemize}

\section{Numerical Results and Analysis}

This section aims to display simulation results to identify the performance of the proposed IR framework and DT-DPG algorithm. The key parameters of the simulation have been listed in Table \ref{SP}.
We adopt the 'Adam' optimizer to train parameters for both actor and critic neural network. For the specific neural network structures, a number of experiments suggested that for the scenario in this paper, the number of nodes for each hidden layer is selected to be $64-128$. The learning rate can be set in a range of $10^{-3}$ to $10^{-5}$. To maximize the overall performance, the discount factor $\gamma = 1$. In order to ensure the stability of the neural network, a sufficiently large memory buffer and a fairly small target update rate $\tau$ have to be guaranteed. Thus, our buffer capacity is $5\times10^4$, while the batch size $e=64$, and $\tau=0.002$. The scale of the OU noise is set as $N(0,0.5)\sim \xi(0,0)$,$N(0,0.4)\sim \xi(1,0)$ for two training phases. Please note that the parameters are only for this simulation, please refer critically when the proposed algorithm is applied to other problems.

\begin{figure}[t]
\centering
\subfigure[Trajectories in NOMA/OMA cases]{
\includegraphics[width=0.6\columnwidth]{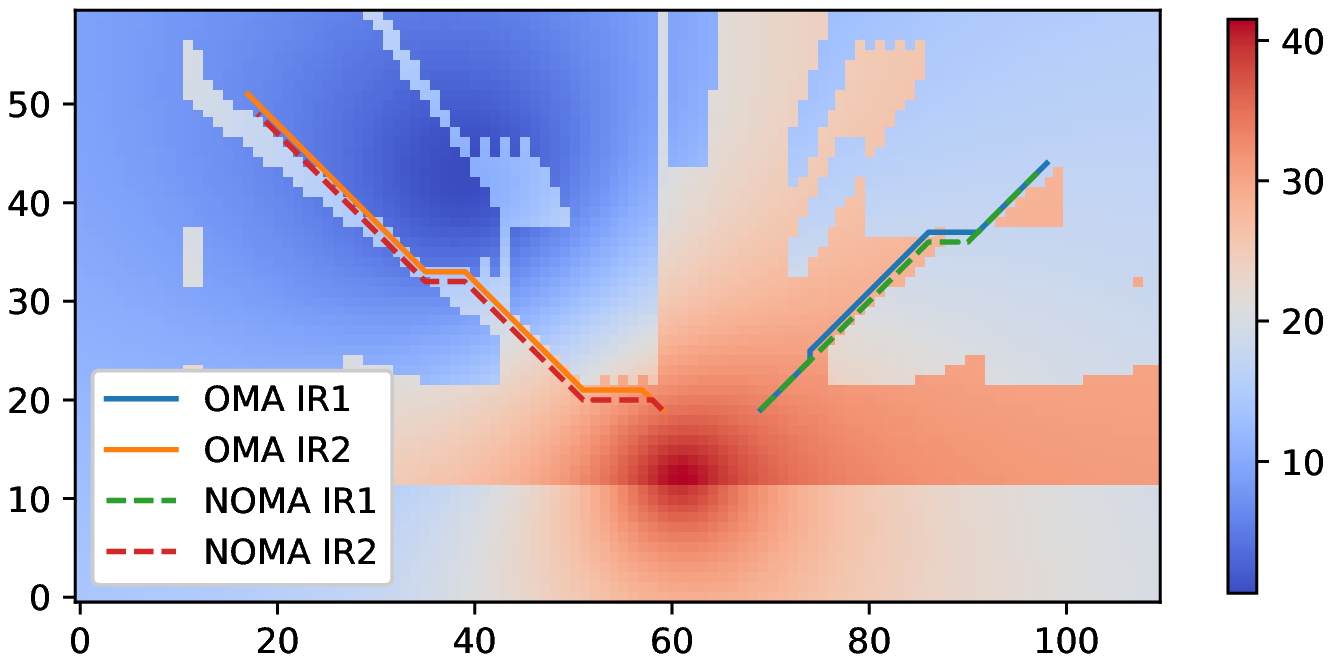}\label{Fig.path1}
}
\quad
\subfigure[Trajectories befroe/after transfer learning]{
\includegraphics[width=0.6\columnwidth]{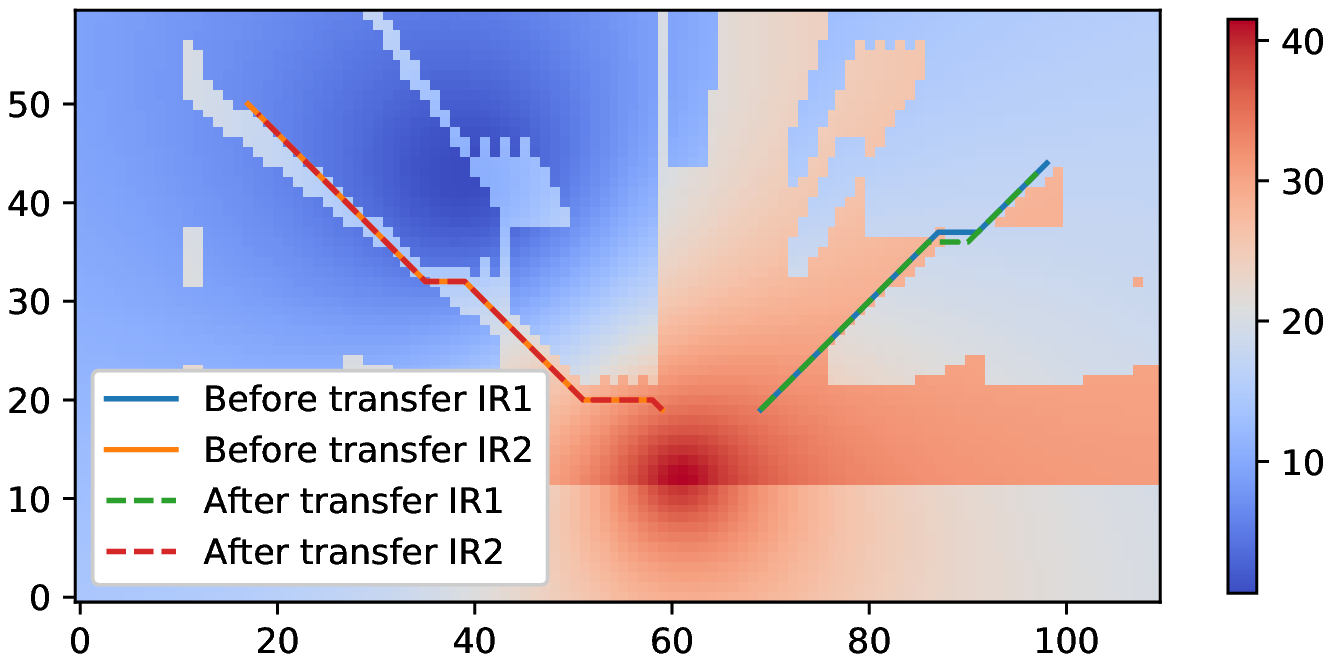}\label{Fig.path2}
}

\caption{IR trajectories in the radio map}\label{Fig.path}
\end{figure}

Fig. \ref{Fig.path} exhibits trajectories of IRs derived from the proposed DT-DPG algorithm and they are displayed in two contrast pairs. The resolution of the radio map is 0.5 meters, and the entire room is divided into $110\times60$ grids. The QoS requirement of each IR is stipulated to be 60kb. Specifically, Fig.\ref{Fig.path1} compares the trajectories of IRs in both OMA and NOMA modes. Understandably, the planned motions keep the IR at a high SINR region as much as possible. In addition, aside from the twists and turns caused by obstacles, the path is also near to the shortest path in terms of spatial distance. Fig.\ref{Fig.path2} compares the trajectories before and after the transfer learning, it can be observed that the MQI training not only optimizes the power allocation but also changes a number of motions.

\begin{table}[t!]

 \caption{Simulation Parameters}\label{SP}
 \centering
 \footnotesize
 \renewcommand\arraystretch{1.5}
 \begin{tabular}{|c|c|c|c|c|c|}
  \hline
  Parameter & Description & Value &   Parameter & Description & Value \\
  \hline
  $f_\text{c}$ & carrier frequency & 2GHz & $U$ & number of IRs & 2  \\
  \hline
  $B$ & bandwidth for each IR & 15 kHz & $P_\text{max}$ & maximum transmitting power & 20 dBm \\
  \hline
  $V_\text{max}$ & maximum speed of IRs & 1 m/s & $\lambda$ & outage penalty coefficient & 1 \\
  \hline
  $h_\text{u}$ & IRs' antenna altitude & 150 m & $h_{\boldsymbol{m}}$ &  altitude of the AP & 2 m \\
  \hline
  $y_\text{max}$ & room length & 30 m & $x_\text{max}$ & room width & 55 m \\
  \hline
  $R_\text{QoS}$& QoS require & 60 kb/s & $\sigma$& AWGN power density & -100 dBm/Hz \\
  \hline
  $\alpha$ & learning rate & $10^{-4}$ & $\gamma$ & discount factor & 1\\
  \hline
  $e$ & batch size & 64 samples & $\tau$ & target update rate & 0.002 \\
  \hline
  $\omega$ & buffer capacity & $5\times10^4$ &  $\xi(0,0)$ & Initial OU noise scale& 0.5\\
  \hline

 \end{tabular}
\end{table}

\begin{figure}[htbp]
\centering
\subfigure[Average reward during destination training]{
\includegraphics[width=0.6\columnwidth]{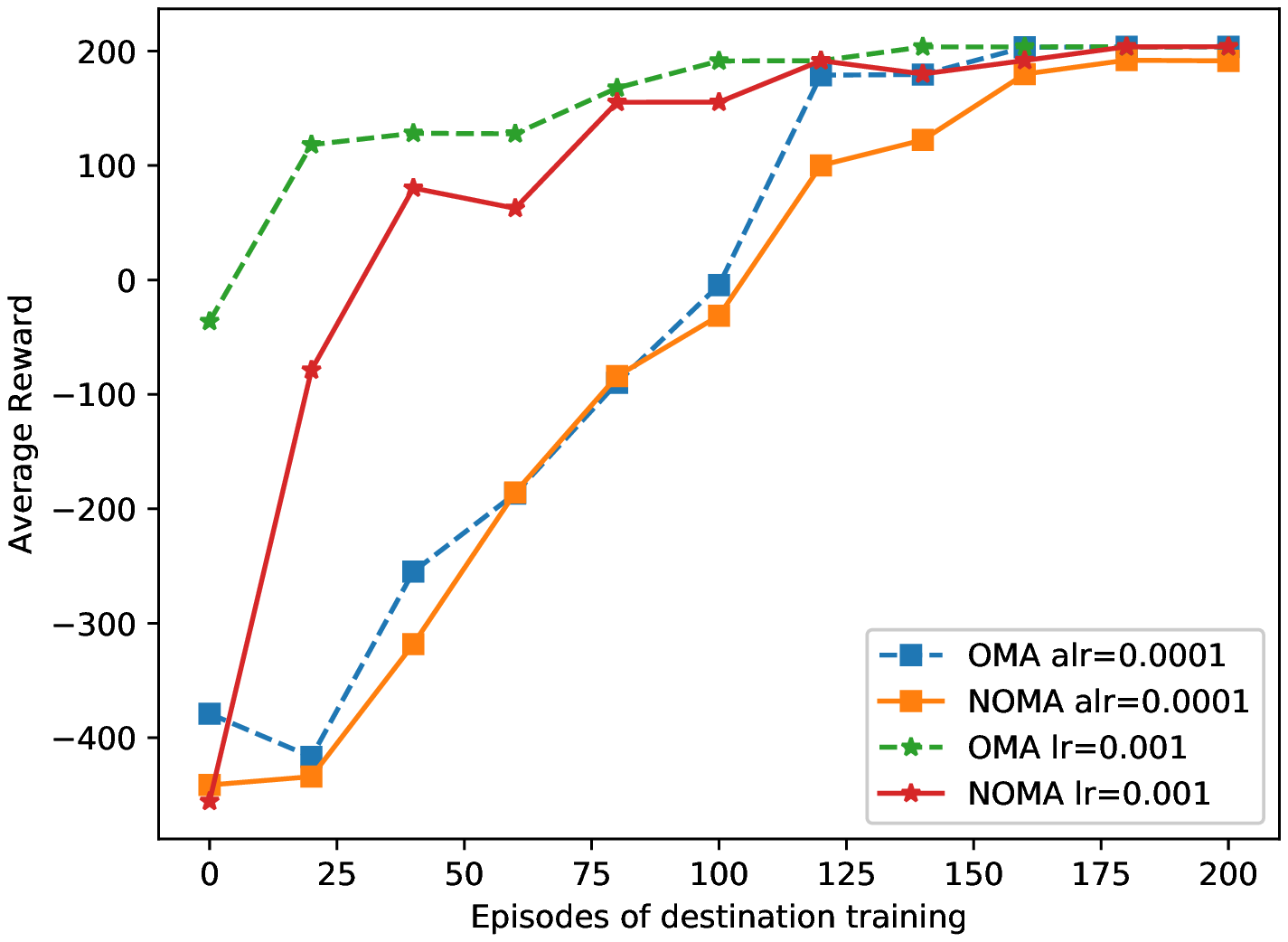}\label{Fig.Reward1}
}
\quad
\subfigure[Average reward during MQI training]{
\includegraphics[width=0.6\columnwidth]{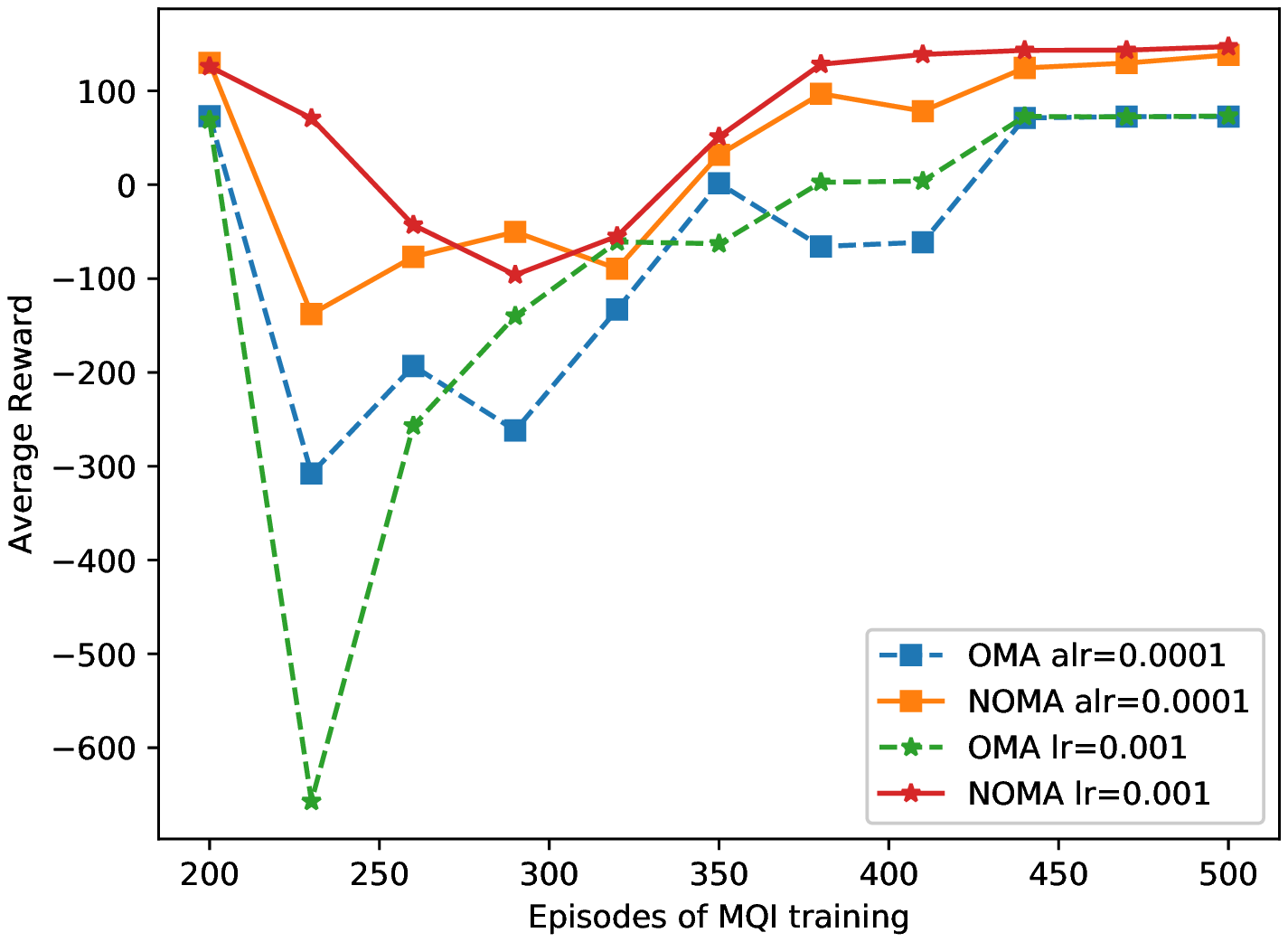}\label{Fig.Reward2}
}

\caption{Average reward of the DT-DPG algorithm in two training phase}\label{Fig.Reward}
\end{figure}

Fig. \ref{Fig.Reward} reveals the accumulated rewards of each episode in two training phases separately. As a premise, it needs to be emphasized that since the different reward functions are employed, comparing the rewards of two training phases is ambiguous and inconsequential. It can be observed that the convergence rate is significantly affected by the learning rate. When the learning rate is 0.0001, the convergence rate is significantly slower but more stable, which does not cause a sharp reward drop after the neural network is transferred in Fig. \ref{Fig.Reward2}. Since the data rate is not considered in the first training phase, the reward is only depended on the path and then all the curves converged into a similar value in Fig. \ref{Fig.Reward1}. However, in the second training phase, the maximized rewards in NOMA and OMA cases have a significant gap, which suggests that with reasonable planning of the path and transmission power, the NOMA framework has more energetic data rate potential than the OMA framework.

\begin{figure}[htbp]
\centering
\includegraphics[width=0.8\columnwidth]{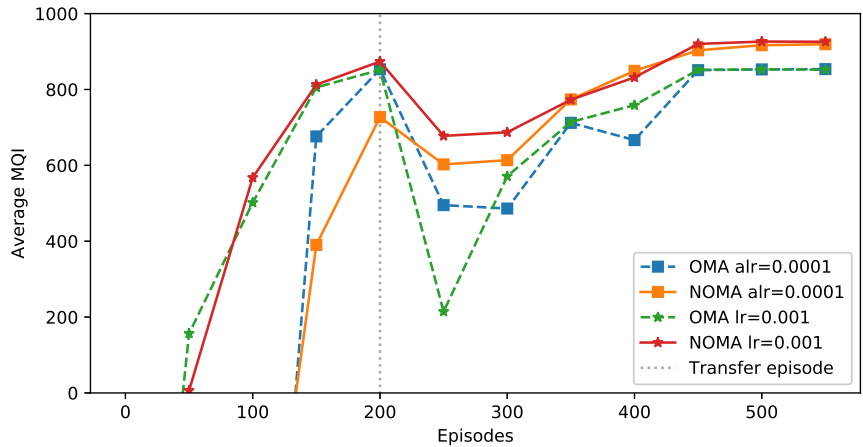}
\caption{Average MQI of the DT-DPG algorithm}\label{Fig.TDDPG}
\end{figure}

\begin{figure}[htbp]
\centering
\includegraphics[width=0.8\columnwidth]{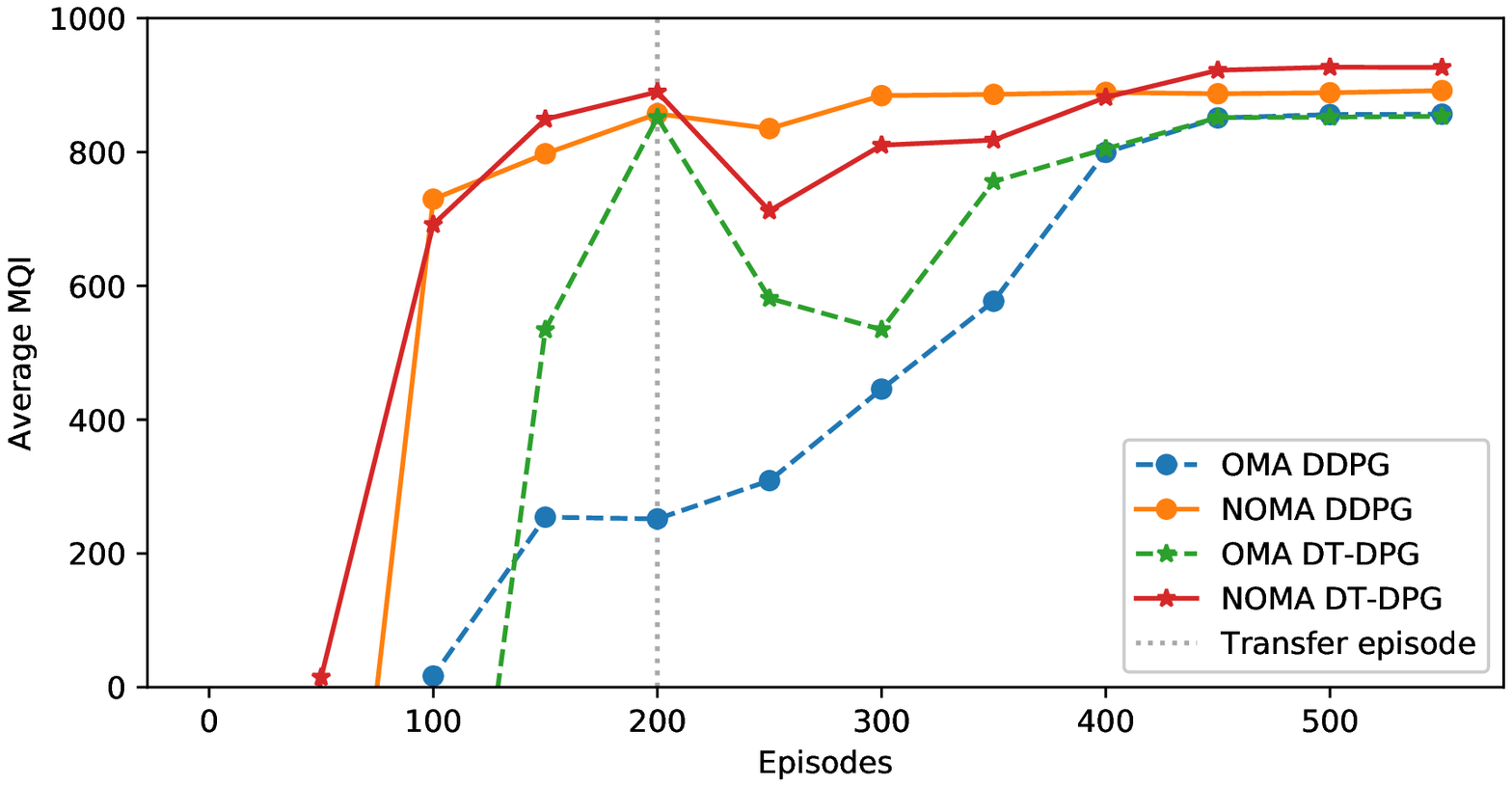}
\caption{Performance of different algorithms}\label{Fig.DDPGvsTDDPG}
\end{figure}

Fig. \ref{Fig.TDDPG} certificates that with an appropriate learning rate range, the proposed DT-DPG algorithm has stable convergence. It can be observed that the convergence of DT-DPG has a special character, namely, multi-stage convergence. In the initial stage of training, the MQI is poor since disorderly actions of the untrained IRs waste mission time on wandering and even fail to reach destinations. After the transfer episode, the curve has a significant depression, which corresponds to the reward dent in Fig. \ref{Fig.Reward2}. The reason for the depression is that the randomness of the action is increased after the transfer, resulting in performance degradation. With the second convergence, the MQI obtained is higher than the first training process, since the MQI training prompts the agent to avoid outages. It needs to be emphasized that when the destination is successfully reached, $T_\text{total} = 1000$ is regarded as the maximum score. Then the gap of dozens of MQI between two convergence physically means that dozens of QoS violations have been avoided, which is a prominent performance improvement though it seems not noteworthy on the curves.

Fig. \ref{Fig.DDPGvsTDDPG} compares the performance of DT-DPG and DDPG algorithms. Excluding the unique convergence curve of DT-DPG, both algorithms are convergent but the DT-DPG algorithm outperforms the DDPG
algorithm at the end of the training. The reason for this phenomenon is that the final reward function of the DT-DPG algorithm is completely consistent with the objective function, but the reward function of the DDPG algorithm is contaminated by the inducing reward (As mentioned in Section 4, the absence of inducing reward can result in an unacceptable training time, since the agent will fall into directionless exploration and are not likely to find the destination).
In addition, it is worth noting that since the transfer learning decomposes multiple tasks, the DT-DPG algorithm is no longer easily trapped in the local optimum as a conventional DDPG algorithm. Our experiments indicate that under the same training constraints, the DDPG algorithm has a success rate of less than 20\% to find the destination while DT-DPG is above 95\%, since the data rate is included in the reward function prematurely, this interferes with choices of the agent during exploration.

\begin{table}[t!]
 \footnotesize
 \renewcommand\arraystretch{1.5}
 \caption{MQI scores in the test episodes}\label{MQI}
 \centering
 \begin{tabular}{|p{2.5cm}<{\centering}||p{5.5cm}<{\centering}||p{3.5cm}<{\centering}||p{1.5cm}<{\centering}|}
  \hline
  MA mode & Approach & Training Episodes & MQI \\
  \hline
        OMA & DT-DPG (Radio map trained) & 500 & 879.5  \\
        \hline
        OMA & DT-DPG (Reality trained)& 500 & 868.5  \\
        \hline
        OMA & DT-DPG (Reality trained) & 700 & 876.2 \\
        \hline

        NOMA & DT-DPG (Radio map trained) & 500 & 929.0  \\
        \hline
        NOMA & DT-DPG (Reality trained)& 500 & 919.0  \\
        \hline
        NOMA & DT-DPG (Reality trained) & 700 & 929.0 \\
        \hline

 \end{tabular}
\end{table}

Table \ref{MQI} shows the average MQI within 20 test episodes and the 'training episodes' represent how many episodes are spent on training the agent. As opposed to the radio map training environment, the test episodes invoke a random fading model to simulate the real environment.
It can be seen that the NOMA scheme has significant improvement over the OMA scheme, due to the preferable user fairness of NOMA techniques. The representation of radio map training is consistent with the training in reality, where the well-trained agents achieved similar performances in both OMA and NOMA cases. Moreover, the radio map training approach improves the training efficiency by costing 200 fewer training episodes than training in reality. The reason is the high randomness of the fading causes interference to the convergence of the neural network, which is excluded by the statistical channel information provided by the radio map.

\section{Conclusions}

This paper proposed an indoor intelligent robot service paradigm and invoked NOMA techniques to improve communication reliability. Based on this paradigm, we formulated a joint path design and power allocation problem to maximize the MQI. We originally proposed a novel reinforcement learning approach named DT-DPG as a solution for the multi-objective problem mixed with destination search, motion planning and power allocation. Moreover, we employed the radio map to train the DT-DPG agent to save hardware costs and improve training efficiency. Our simulation results confirmed that the NOMA scheme improves the communication reliability of IRs compare to the OMA scheme. Furthermore, the trajectories derived from the DT-DPG algorithm have higher MQI and a larger probability to reach the mission destination than the conventional DDPG algorithm. At last, the radio map is a qualified virtual training environment for the RL agent, which can also save training time and hardware costs.


\renewcommand{\baselinestretch}{1.5}
\bibliography{Robotref_abb}

\begin{thebibliography}{10}
\providecommand{\url}[1]{#1}
\csname url@samestyle\endcsname
\providecommand{\newblock}{\relax}
\providecommand{\bibinfo}[2]{#2}
\providecommand{\BIBentrySTDinterwordspacing}{\spaceskip=0pt\relax}
\providecommand{\BIBentryALTinterwordstretchfactor}{4}
\providecommand{\BIBentryALTinterwordspacing}{\spaceskip=\fontdimen2\font plus
\BIBentryALTinterwordstretchfactor\fontdimen3\font minus
  \fontdimen4\font\relax}
\providecommand{\BIBforeignlanguage}[2]{{%
\expandafter\ifx\csname l@#1\endcsname\relax
\typeout{** WARNING: IEEEtran.bst: No hyphenation pattern has been}%
\typeout{** loaded for the language `#1'. Using the pattern for}%
\typeout{** the default language instead.}%
\else
\language=\csname l@#1\endcsname
\fi
#2}}
\providecommand{\BIBdecl}{\relax}
\BIBdecl

\bibitem{tan2019one}
X.~Z. Tan, S.~Reig, E.~J. Carter, and A.~Steinfeld, ``From one to another: how
  robot-robot interaction affects users' perceptions following a transition
  between robots,'' in \emph{2019 14th ACM/IEEE Int. Conf. Hum.-Rob.
  Interact.}\hskip 1em plus 0.5em minus 0.4em\relax IEEE, 2019, pp. 114--122.

\bibitem{lindhe2012communication}
M.~M. Lindhe, ``Communication-aware motion planning for mobile robots,'' Ph.D.
  dissertation, PhD Dissertation, KTH, Stockholm, 2012.

\bibitem{galambos2020cloud}
P.~Galambos, ``Cloud, fog, and mist computing: Advanced robot applications,''
  \emph{IEEE Syst., Man, and Cybern. Mag.}, vol.~6, no.~1, pp. 41--45, 2020.

\bibitem{NOMA5G.yuanwei}
Y.~{Liu}, Z.~{Qin}, M.~{Elkashlan}, Z.~{Ding}, A.~{Nallanathan}, and
  L.~{Hanzo}, ``Nonorthogonal multiple access for {5G} and beyond,''
  \emph{Proc. IEEE}, vol. 105, no.~12, pp. 2347--2381, 2017.

\bibitem{zhang2013robot}
Y.~Zhang, D.-W. Gong, and J.-H. Zhang, ``Robot path planning in uncertain
  environment using multi-objective particle swarm optimization,''
  \emph{Neurocomputing}, vol. 103, pp. 172--185, 2013.

\bibitem{bhattacharjee2011multi}
P.~Bhattacharjee, P.~Rakshit, I.~Goswami, A.~Konar, and A.~K. Nagar,
  ``Multi-robot path-planning using artificial bee colony optimization
  algorithm,'' in \emph{2011 NaBIC}.\hskip 1em plus 0.5em minus 0.4em\relax
  IEEE, 2011, pp. 219--224.

\bibitem{li2012efficient}
G.~Li, A.~Yamashita, H.~Asama, and Y.~Tamura, ``An efficient improved
  artificial potential field based regression search method for robot path
  planning,'' in \emph{2012 IEEE ICMA}.\hskip 1em plus 0.5em minus 0.4em\relax
  IEEE, 2012, pp. 1227--1232.

\bibitem{ghaffarkhah2012optimal}
A.~Ghaffarkhah and Y.~Mostofi, ``Optimal motion and communication for
  persistent information collection using a mobile robot,'' in \emph{2012 IEEE
  Globecom Workshops}.\hskip 1em plus 0.5em minus 0.4em\relax IEEE, 2012, pp.
  1532--1537.

\bibitem{ghaffarkhah2011communication}
A.~Ghaffarkhah and Y.~Mostofi, ``Communication-aware motion planning in mobile
  networks,'' \emph{IEEE Trans. Automat. Contr.}, vol.~56, no.~10, pp.
  2478--2485, 2011.

\bibitem{huang2019online}
B.~Huang, Z.~Xu, B.~Jia, and G.~Mao, ``An online radio map update scheme for
  wifi fingerprint-based localization,'' \emph{IEEE Internet Things J.},
  vol.~6, no.~4, pp. 6909--6918, 2019.

\bibitem{utkovski2019learning}
Z.~Utkovski, P.~Agostini, M.~Frey, I.~Bjelakovic, and S.~Stanczak, ``Learning
  radio maps for physical-layer security in the radio access,'' in \emph{2019
  IEEE 20th SPAWC}.\hskip 1em plus 0.5em minus 0.4em\relax IEEE, 2019, pp.
  1--5.

\bibitem{zhang2019radio}
S.~Zhang and R.~Zhang, ``Radio map based 3d path planning for
  cellular-connected {UAV},'' \emph{arXiv}, pp. arXiv--1912.00\,021, 2019.

\bibitem{liu2020machine}
X.~Liu, Y.~Liu, and Y.~Chen, ``Machine learning empowered trajectory and
  passive beamforming design in {UAV-RIS} wireless networks,'' \emph{arXiv
  preprint arXiv:2010.02749}, 2020.

\bibitem{mu2020intelligent}
X.~Mu, Y.~Liu, L.~Guo, J.~Lin, and R.~Schober, ``Intelligent reflecting surface
  enhanced indoor robot path planning: A radio map based approach,''
  \emph{arXiv preprint arXiv:2009.12804}, 2020.

\bibitem{budhiraja2019tactile}
I.~Budhiraja, S.~Tyagi, S.~Tanwar, N.~Kumar, and J.~J. Rodrigues, ``Tactile
  internet for smart communities in {5G}: An insight for {NOMA}-based
  solutions,'' \emph{IEEE Trans. Ind. Inform.}, vol.~15, no.~5, pp. 3104--3112,
  2019.

\bibitem{shirvanimoghaddam2017massive}
M.~Shirvanimoghaddam, M.~Dohler, and S.~J. Johnson, ``Massive non-orthogonal
  multiple access for cellular {IoT}: Potentials and limitations,'' \emph{IEEE
  Commun. Mag.}, vol.~55, no.~9, pp. 55--61, 2017.

\bibitem{Yuanwei.1}
Y.~{Liu}, Z.~{Qin}, Y.~{Cai}, Y.~{Gao}, G.~Y. {Li}, and A.~{Nallanathan},
  ``{UAV} communications based on non-orthogonal multiple access,'' \emph{IEEE
  Wireless Commun.}, vol.~26, no.~1, pp. 52--57, 2019.

\bibitem{zhang2017downlink}
Z.~Zhang, H.~Sun, and R.~Q. Hu, ``Downlink and uplink non-orthogonal multiple
  access in a dense wireless network,'' \emph{IEEE J. Sel. Areas Commun.},
  vol.~35, no.~12, pp. 2771--2784, 2017.

\bibitem{zhai2019delay}
D.~Zhai, R.~Zhang, L.~Cai, and F.~R. Yu, ``Delay minimization for massive
  {Internet of Things} with non-orthogonal multiple access,'' \emph{IEEE J.
  Sel. Top. Signal Process.}, vol.~13, no.~3, pp. 553--566, 2019.

\bibitem{yang2016general}
Z.~Yang, Z.~Ding, P.~Fan, and N.~Al-Dhahir, ``A general power allocation scheme
  to guarantee quality of service in downlink and uplink {NOMA} systems,''
  \emph{IEEE Trans. Wirel. Commun.}, vol.~15, no.~11, pp. 7244--7257, 2016.

\bibitem{zhai2018energy}
D.~Zhai, R.~Zhang, L.~Cai, B.~Li, and Y.~Jiang, ``Energy-efficient user
  scheduling and power allocation for {NOMA}-based wireless networks with
  massive iot devices,'' \emph{IEEE Internet Things J.}, vol.~5, no.~3, pp.
  1857--1868, 2018.

\bibitem{zhang2020continuous}
K.~Zhang, S.~McLeod, M.~Lee, and J.~Xiao, ``Continuous reinforcement learning
  to adapt multi-objective optimization online for robot motion,'' \emph{Int J
  Adv Robot Syst.}, vol.~17, no.~2, 2020.

\bibitem{lobos2018visual}
K.~Lobos-Tsunekawa, F.~Leiva, and J.~Ruiz-del Solar, ``Visual navigation for
  biped humanoid robots using deep reinforcement learning,'' \emph{IEEE Robot.
  Autom. Lett.}, vol.~3, no.~4, pp. 3247--3254, 2018.

\bibitem{kakish2020using}
Z.~M. Kakish, K.~Elamvazhuthi, and S.~Berman, ``Using reinforcement learning to
  herd a robotic swarm to a target distribution,'' \emph{arXiv preprint
  arXiv:2006.15807}, 2020.

\bibitem{lakshmanan2020complete}
A.~K. Lakshmanan, R.~E. Mohan, B.~Ramalingam, A.~V. Le, P.~Veerajagadeshwar,
  K.~Tiwari, and M.~Ilyas, ``Complete coverage path planning using
  reinforcement learning for tetromino based cleaning and maintenance robot,''
  \emph{Autom. Constr.}, vol. 112, p. 103078, 2020.

\bibitem{james2019online}
J.~James, W.~Yu, and J.~Gu, ``Online vehicle routing with neural combinatorial
  optimization and deep reinforcement learning,'' \emph{IEEE trans. Intell.
  Transp. Syst.}, vol.~20, no.~10, pp. 3806--3817, 2019.

\bibitem{liu2020enhancing}
X.~Liu, Y.~Liu, Y.~Chen, and L.~Hanzo, ``Enhancing the fuel-economy of
  {V2I}-assisted autonomous driving: A reinforcement learning approach,''
  \emph{IEEE Trans. Veh. Technol.}, vol.~69, no.~8, pp. 8329--8342, 2020.

\bibitem{liu2018energy}
C.~H. Liu, Z.~Chen, J.~Tang, J.~Xu, and C.~Piao, ``Energy-efficient {UAV}
  control for effective and fair communication coverage: A deep reinforcement
  learning approach,'' \emph{IEEE J. Sel. Areas Commun}, vol.~36, no.~9, pp.
  2059--2070, 2018.

\bibitem{liu2019trajectory}
X.~Liu, Y.~Liu, Y.~Chen, and L.~Hanzo, ``Trajectory design and power control
  for multi-{UAV} assisted wireless networks: A machine learning approach,''
  \emph{IEEE Trans. Veh. Technol.}, vol.~68, no.~8, pp. 7957--7969, 2019.

\bibitem{chen2019liquid}
M.~Chen, W.~Saad, and C.~Yin, ``Liquid state machine learning for resource and
  cache management in lte-u unmanned aerial vehicle (uav) networks,''
  \emph{IEEE Trans. Wirel. Commun.}, vol.~18, no.~3, pp. 1504--1517, 2019.

\bibitem{9140367}
Y.~{Wang}, M.~{Chen}, Z.~{Yang}, T.~{Luo}, and W.~{Saad}, ``Deep learning for
  optimal deployment of {UAV}s with visible light communications,'' \emph{IEEE
  Trans. Wirel. Commun.}, vol.~19, no.~11, pp. 7049--7063, 2020.

\bibitem{liu2020}
X.~Liu, M.~Chen, Y.~Liu, Y.~Chen, S.~Cui, and L.~Hanzo, ``Artificial
  intelligence aided next-generation networks relying on {UAV}s,'' \emph{arXiv
  preprint arXiv:2001.11958}, 2020.

\bibitem{qian2019survey}
Y.~Qian, J.~Wu, R.~Wang, F.~Zhu, and W.~Zhang, ``Survey on reinforcement
  learning applications in communication networks,'' \emph{J. Commun. Info.
  Netw.}, 2019.

\bibitem{jingjingwang1}
J.~{Wang}, S.~{Guan}, C.~{Jiang}, D.~{Alanis}, Y.~{Ren}, and L.~{Hanzo},
  ``Network association in machine-learning aided cognitive radar and
  communication co-design,'' \emph{IEEE IEEE J. Sel. Areas Commun.}, vol.~37,
  no.~10, pp. 2322--2336, 2019.

\bibitem{yang2020deep}
H.~Yang, Z.~Xiong, J.~Zhao, D.~Niyato, L.~Xiao, and Q.~Wu, ``Deep reinforcement
  learning based intelligent reflecting surface for secure wireless
  communications,'' \emph{IEEE Trans. Wirel. Commun.}, 2020.

\bibitem{zhao2019deep}
N.~Zhao, Y.-C. Liang, D.~Niyato, Y.~Pei, M.~Wu, and Y.~Jiang, ``Deep
  reinforcement learning for user association and resource allocation in
  heterogeneous cellular networks,'' \emph{IEEE Trans. Wirel. Commun.},
  vol.~18, no.~11, pp. 5141--5152, 2019.

\bibitem{he2019joint}
C.~He, Y.~Hu, Y.~Chen, and B.~Zeng, ``Joint power allocation and channel
  assignment for {NOMA} with deep reinforcement learning,'' \emph{IEEE J. Sel.
  Areas Commun}, vol.~37, no.~10, pp. 2200--2210, 2019.

\bibitem{chen2019artificial}
M.~Chen, U.~Challita, W.~Saad, C.~Yin, and M.~Debbah, ``Artificial neural
  networks-based machine learning for wireless networks: A tutorial,''
  \emph{IEEE Commun. Surv. Tutor.}, vol.~21, no.~4, pp. 3039--3071, 2019.

\bibitem{jingjingwang2}
J.~{Wang}, C.~{Jiang}, H.~{Zhang}, Y.~{Ren}, K.~C. {Chen}, and L.~{Hanzo},
  ``Thirty years of machine learning: The road to pareto-optimal wireless
  networks,'' \emph{IEEE Commun. Surv. Tutor.}, vol.~22, no.~3, pp. 1472--1514,
  2020.

\bibitem{ITUR}
R.~I.-R. P.1238-10, ``Propagation data and prediction methods for the planning
  of indoor radiocommunication systems and radio local area networks in the
  frequency range 300 {MH}z to 450 {GH}z,'' 2019.

\bibitem{ITUR2}
R.~I.-R. M.2135-1, ``Guidelines for evaluation of radio interface technologies
  for {IMT}-advanced,'' 2009.

\bibitem{liu2017non}
Y.~Liu, Z.~Qin, M.~Elkashlan, Z.~Ding, A.~Nallanathan, and L.~Hanzo,
  ``Non-orthogonal multiple access for {5G} and beyond,'' \emph{Proc. IEEE},
  vol. 105, no.~12, pp. 2347--2381, 2017.

\bibitem{cui.signal}
J.~{Cui}, Y.~{Liu}, Z.~{Ding}, P.~{Fan}, and A.~{Nallanathan}, ``Optimal user
  scheduling and power allocation for millimeter wave {NOMA} systems,''
  \emph{IEEE Trans. Wireless Commun.}, vol.~17, no.~3, pp. 1502--1517, 2018.

\bibitem{bi2019engineering}
S.~Bi, J.~Lyu, Z.~Ding, and R.~Zhang, ``Engineering radio maps for wireless
  resource management,'' \emph{IEEE Wireless Commun.}, vol.~26, no.~2, pp.
  133--141, 2019.

\bibitem{zou2017winips}
H.~Zou, M.~Jin, H.~Jiang, L.~Xie, and C.~J. Spanos, ``Winips: {WiFi}-based
  non-intrusive indoor positioning system with online radio map construction
  and adaptation,'' \emph{IEEE Trans. on Wireless Commun.}, vol.~16, no.~12,
  pp. 8118--8130, 2017.

\bibitem{majercik2018ray}
A.~Majercik, C.~Crassin, P.~Shirley, and M.~McGuire, ``A ray-box intersection
  algorithm and efficient dynamic voxel rendering,'' \emph{Journal of Computer
  Graphics Techniques Vol}, vol.~7, no.~3, 2018.

\bibitem{matheron2019problem}
G.~Matheron, N.~Perrin, and O.~Sigaud, ``The problem with {DDPG}: understanding
  failures in deterministic environments with sparse rewards,'' \emph{arXiv
  preprint arXiv:1911.11679}, 2019.

\bibitem{li2019robust}
S.~Li, Y.~Wu, X.~Cui, H.~Dong, F.~Fang, and S.~Russell, ``Robust multi-agent
  reinforcement learning via minimax deep deterministic policy gradient,'' in
  \emph{Proceedings of the AAAI Conference on Artificial Intelligence},
  vol.~33, 2019, pp. 4213--4220.

\bibitem{lillicrap2015continuous}
T.~P. Lillicrap, J.~J. Hunt, A.~Pritzel, N.~Heess, T.~Erez, Y.~Tassa,
  D.~Silver, and D.~Wierstra, ``Continuous control with deep reinforcement
  learning,'' \emph{arXiv preprint arXiv:1509.02971}, 2015.

\bibitem{colas2018gep}
C.~Colas, O.~Sigaud, and P.-Y. Oudeyer, ``Gep-pg: Decoupling exploration and
  exploitation in deep reinforcement learning algorithms,'' \emph{arXiv
  preprint arXiv:1802.05054}, 2018.

\bibitem{qiu2019deep}
C.~Qiu, Y.~Hu, Y.~Chen, and B.~Zeng, ``Deep deterministic policy gradient
  ({DDPG})-based energy harvesting wireless communications,'' \emph{IEEE
  Internet Things J.}, vol.~6, no.~5, pp. 8577--8588, 2019.

\end{thebibliography}
\bibliographystyle{IEEEtran}

\end{document}